\documentclass[11pt]{article}

\usepackage{latexsym,mathrsfs}
\usepackage{amsmath,amssymb}
\usepackage{amsthm,enumerate,verbatim}
\usepackage{amsfonts}
\usepackage{graphicx}
\usepackage{algorithm}
\usepackage{algorithmic}
\usepackage{url}

\newtheorem*{definition*}{Definition}

\DeclareMathOperator{\argmax}{argmax} 
\DeclareMathOperator{\argmin}{argmin} 
 
\DeclareMathOperator{\rank}{rank}

\DeclareMathOperator{\diag}{diag}

\title{The Why and How of Nonnegative Matrix Factorization}
\date{}

\author{Nicolas Gillis \\ % \thanks{xxx} 
Department of Mathematics and Operational Research \\ 
Facult\'e Polytechnique, Universit\'e de Mons \\ 
Rue de Houdain 9, 7000 Mons, Belgium\\
 nicolas.gillis@umons.ac.be  
}

\begin{document}

\maketitle

\begin{abstract} 
Nonnegative matrix factorization (NMF) has become a widely used tool for the analysis of high-dimensional data as it automatically extracts sparse and meaningful features from a set of nonnegative data vectors. 
We first illustrate this property of NMF on three applications, in image processing, text mining and hyperspectral imaging --this is the why. 
Then we address the problem of solving NMF, which is NP-hard in general. 
We review some standard NMF algorithms, and also present a recent subclass of NMF problems, referred to as near-separable NMF, that can be solved efficiently (that is, in polynomial time), even in the presence of noise --this is the how. 
Finally, we briefly describe some problems in mathematics and computer science closely related to NMF via the nonnegative rank. 
\end{abstract}


\section{Introduction}

Linear dimensionality reduction (LDR) techniques are a key tool in data analysis, and are widely used for example for compression, visualization, feature selection and noise filtering.  
Given a set of data points $x_j \in \mathbb{R}^p$ for $1 \leq j \leq n$ and a dimension $r < \min(p,n)$, LDR amounts to computing a set of $r$ basis elements $w_k \in \mathbb{R}^p$ for $1 \leq k \leq r$ such that the linear space spanned by the $w_k$'s approximates the data points as closely as possible, that is, such that we have for all $j$  
\begin{equation} \label{ldr}
x_j \approx \sum_{k=1}^r w_k h_j(k), \qquad \text{ for some weights $h_j \in \mathbb{R}^r$}. 
\end{equation}
In other words, the $p$-dimensional data points are represented in a $r$-dimensional linear subspace spanned by the basis elements $w_k$'s and whose coordinates are given by the vectors $h_j$'s. LDR is equivalent to low-rank matrix approximation: in fact, constructing 
\begin{itemize}

\item the matrix $X \in \mathbb{R}^{p \times n}$ such that each column is a data point, 
that is, 
$X(:,j) = x_j$ for $1 \leq j \leq n$, 

\item the matrix $W \in \mathbb{R}^{p \times r}$ such that each column is a basis element, 
that is, 
$W(:,k) = w_k$ for $1 \leq k \leq r$, and

\item the matrix $H \in \mathbb{R}^{r \times n}$  such that each column of $H$ gives the coordinates of a data point $X(:,j)$ in the basis $W$, 
that is, 
\mbox{$H(:,j) = h_j$} for $1 \leq j \leq n$,  

\end{itemize}
the above LDR model \eqref{ldr} is equivalent to $X \approx WH$, that is, to approximate the data matrix $X$ with a low-rank matrix $WH$. 

A first key aspect of LDR is the choice of the measure to assess the quality of the approximation. It should be chosen depending on the \textit{noise model}. 
The most widely used measure is the Frobenius norm of the error, that is, $||X-WH||_F^2 = \sum_{i,j} (X-WH)_{ij}^2$.
The reason for the popularity of the Frobenius norm is two-fold. First, it implicitly assumes the noise $N$ present in the matrix $X = WH + N$ to be Gaussian, which is reasonable in many practical situations (see also the introduction of Section~\ref{how}). 
%it implicitly assumes that each entry of the noise $N$ added to the matrix $X = WH + N$ is Gaussian which is reasonable in many practical situations. 
%implicitly assuming that the random variation in the tensor data follows a Gaussian distribution
Second, an optimal approximation can be computed efficiently through the truncated singular value decomposition (SVD); see~\cite{GV96} and the references therein. Note that the SVD is equivalent to principal component analysis (PCA) after mean centering of the data points (that is, after shifting all data points so that their mean is on the origin). 

A second key aspect of LDR is the assumption on the structure of the factors $W$ and $H$. 
The truncated SVD and PCA do not make any assumption on $W$ and $H$. 
For example, assuming independence of the columns of $W$ leads to independent component analysis (ICA)~\cite{C94}, 
or assuming sparsity of $W$ (and/or $H$) leads to sparse low-rank matrix decompositions, such as sparse PCA~\cite{AE07}. 
%In this paper, we focus on the model where both factors are component-wise nonnegative, leading to nonnegative matrix factorization (NMF). 
%Three popular are: 
%\begin{itemize} 
%\item Independant component analysis (ICA) where the columns of $W$ are assumed to be independent. 
%\item , where the columns of  are assumed to be sparse.  
%%\item robust PCA? 
%\item Nonnegative matrix factorization (NMF) 
%\end{itemize} 
%\subsection{
Nonnegative matrix factorization (NMF) is an LDR where both the basis elements $w_k$'s and the weights $h_j$'s are assumed to be \emph{component-wise nonnegative}. Hence  NMF aims at decomposing a given nonnegative data matrix $X$ as $X \approx WH$ where $W \geq 0$ and $H \geq 0$ (meaning that $W$ and $H$ are component-wise nonnegative). 
NMF was first introduced in 1994 by Paatero and Tapper~\cite{PT94} and gathered more and more interest after an article by Lee and Seung~\cite{LS99} in 1999. 

In this paper, we explain \emph{why} NMF has been so popular in different data mining applications, and \emph{how} one can compute NMF's. The aim of this paper is not to give a comprehensive overview of all NMF applications and algorithms 
--and we apologize for not mentioning many relevant contributions-- %and for being biased toward my own expertise-- 
but rather to serve as an introduction to NMF, describing three applications and several standard algorithms.

%\subsection{Notation} 

\section{The Why -- NMF Generates Sparse and Meaningful Features} 

The reason why NMF has become so popular is because of its ability to automatically extract sparse and easily interpretable factors. 
In this section, we illustrate this property of NMF through three applications, in image processing, text mining and hyperspectral imaging. 
Other applications include air emission control~\cite{PT94}, %image processing~\cite{LS99}, 
computational biology~\cite{D08}, blind source separation~\cite{CMCW08}, 
single-channel source separation~\cite{AL11}, 
clustering~\cite{DHS05}, 
music analysis~\cite{FCB09}, collaborative filtering~\cite{Vikas10}, and 
community detection~\cite{WL11}.

\subsection{Image Processing -- Facial Feature Extraction}  \label{image}

\begin{figure}[ht!]
\begin{center}
\[
\hspace{-0.5cm} 
\underbrace{X(:,j)}_{\text{\begin{tabular}{c} $j$th facial image \vspace{0.5cm}\\ \includegraphics[height=1.5cm]{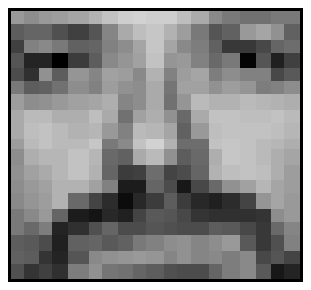} \end{tabular}}} 
\approx   \quad 
\sum_{k=1}^r  
\underbrace{W(:,k)}_{\text{\begin{tabular}{c} facial features \vspace{0.2cm} \\ \includegraphics[height=4cm]{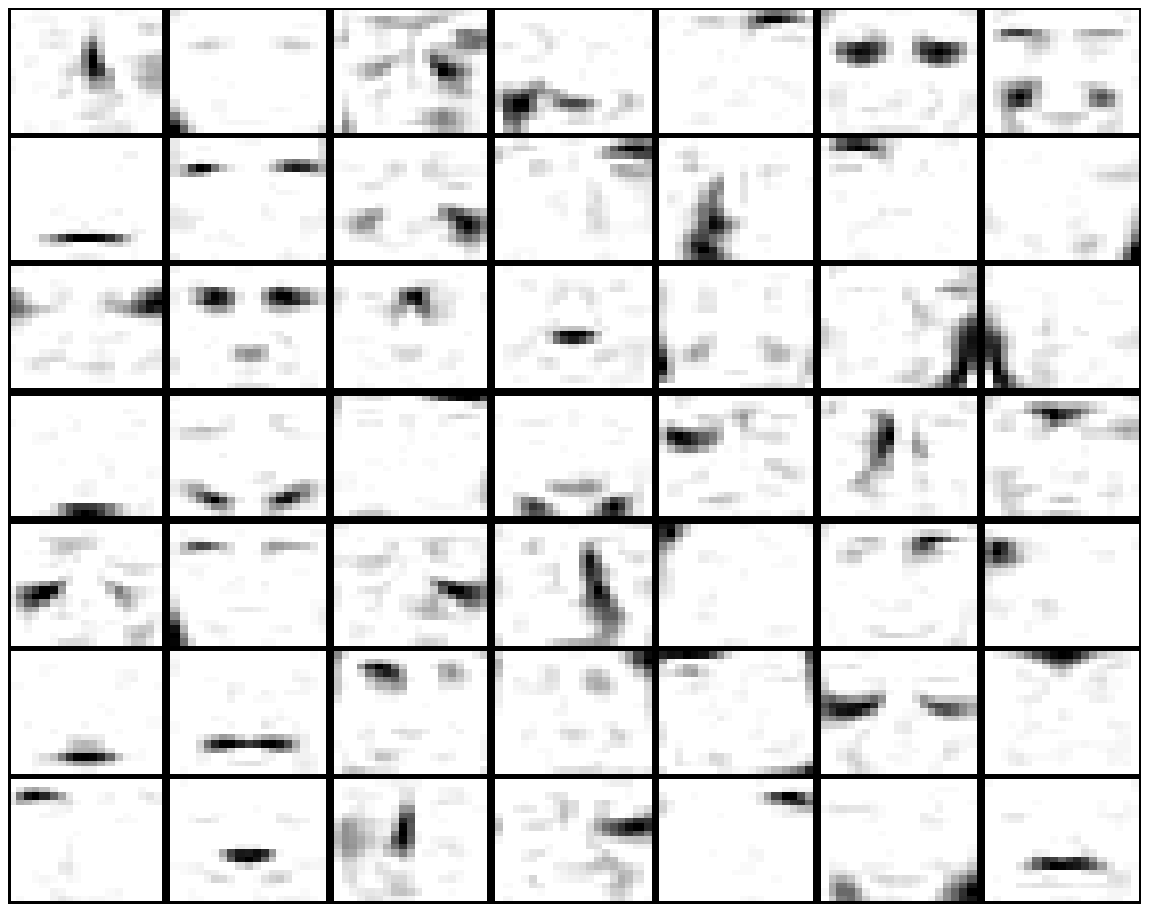} \end{tabular}}}   
\underbrace{H(k,j)}_{\text{\begin{tabular}{c} importance of features \\ in $j$th image \vspace{0.2cm} \\  \includegraphics[height=2cm]{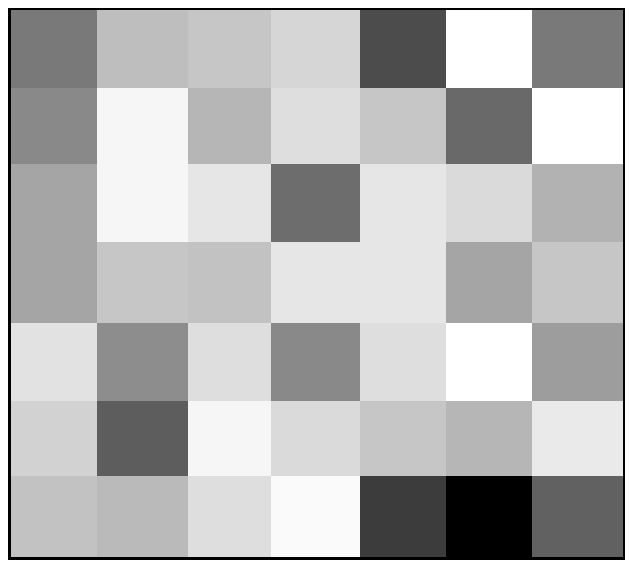} \end{tabular}}} 
= 
\underbrace{WH(:,j)}_{\text{\begin{tabular}{c} approximation \\ of $j$th image \vspace{0.2cm} \\ \includegraphics[height=1.5cm]{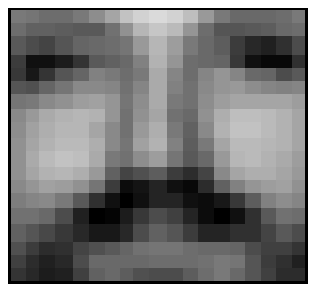} \end{tabular}}}.  
\]
\caption{\small Decomposition of the CBCL face database, MIT Center For Biological and Computation Learning (2429 gray-level 19-by-19 pixels images) using $r=49$ as in~\cite{LS99}.} 
\label{cbcl}
\end{center}
\end{figure}

Let each column of the data matrix $X \in \mathbb{R}^{p \times n}_+$ be a vectorized gray-level image of a face, with the $(i,j)$th entry of matrix $X$ being the intensity of the $i$th pixel in the $j$th face.
NMF generates two factors $(W,H)$ so that each image $X(:,j)$ is approximated using a linear combination of the columns of $W$; see Equation~\eqref{ldr}, and Figure~\ref{cbcl} for an illustration.  
Since $W$ is nonnegative, the columns of $W$ can be interpreted as images (that is, vectors of pixel intensities) which we refer to as the basis images. 
As the weights in the linear combinations are nonnegative ($H \geq 0$), these basis images can only be summed up to reconstruct each original image. 
Moreover, the large number of images in the data set must be reconstructed approximately with only a few basis images (in fact, $r$ is in general much smaller than $n$), hence 
%the number of basis images is much smaller than the number of images in the data set , 
the latter should be localized features (hence sparse) found simultaneously in several images. 
%Therefore, NMF naturally extracts localized. 
In the case of facial images, the basis images are features such as eyes, noses, mustaches, and lips (see Figure~\ref{cbcl}) while the columns of $H$ indicate which feature is present in which image (see also \cite{LS99, GV02}). 

A potential application of NMF is in face recognition. It has for example been observed that NMF is more robust to occlusion than PCA (which generates dense factors): in fact, if a new occluded face (e.g., with sun glasses) has to be mapped into the NMF basis, the non-occluded parts (e.g., the mustache or the lips) can still be well approximated~\cite{GV02}.

\subsection{Text Mining -- Topic Recovery and Document Classification} \label{text}

Let each column of the nonnegative data matrix $X$ correspond to a document and each row to a word. 
The $(i,j)$th entry of the matrix $X$ could for example be equal to the number of times the $i$th word appears in the $j$th document in which case each column of $X$ is the vector of word counts of a document; in practice, more sophisticated constructions are used, e.g., the term frequency - inverse document frequency (tf-idf).  
This is the so-called bag-of-words model: each document is associated with a set of words with different weights, while the ordering of the words in the documents is not taken into account (see, e.g., the survey \cite{Blei12} for a discussion).  
Note that such a matrix $X$ is in general rather sparse as most documents only use a small subset of the dictionary. 
Given such a matrix $X$ and a factorization rank $r$, NMF generates two factors $(W,H)$ such that, for all $1 \leq j \leq n$, we have 
\[
\underbrace{X(:,j)}_{\text{$j$th document}} 
\approx \quad  
\sum_{k=1}^r \quad 
\underbrace{W(:,k)}_{\text{$k$th topic}} \quad  
\underbrace{H(k,j)}_{\text{\begin{tabular}{c} importance of $k$th topic \\ in $j$th document \end{tabular}}}, 
\qquad \text{ with $W \geq 0$ and  $H \geq 0$. }
\]
 This decomposition can be interpreted as follows (see, also, e.g., \cite{LS99, SBPP06, Ar13}):  
%This NMF of $X$ can be interpreted as follows  %, that is, each document $X(:,j)$ is approximated using a linear combination of the columns of $W$. 
%Because of the nonnegativity constraints, he decomposition can be interpreted as follows: 
\begin{itemize}

\item Because $W$ is nonnegative, each column of $W$ can be interpreted as a document, that is, as a bag of words.  

\item  Because the weights in the linear combinations are nonnegative ($H \geq 0$), one can only take the union of the sets of words defined by the columns of $W$ to reconstruct all the original documents.

\item Moreover, because the number of documents in the data set is much larger than the number of basis elements (that is, the number of columns of $W$), 
the latter should be set of words found simultaneously in several documents. 
%(In fact, in practice, $r$ is usually much smaller than $n$, and if each column of $W$ was only used to approximate a few columns of $X$, the approximation would be rather poor.) 
%(in fact, if they are present in only a few documents,. 
Hence the basis elements can be interpreted as \emph{topics}, that is, set of words found simultaneously in different documents, while the weights in the linear combinations (that is, the matrix $H$) assign the documents to the different topics, that is, identify which document discusses which topic. 

\end{itemize} 

Therefore, given a set of documents, 
\emph{NMF identifies  topics and simultaneously classifies the documents among these different topics.} 
Note that NMF is closely related to existing topic models, in particular probabilistic latent semantic analysis and indexing (PLSA and PLSI)~\cite{GG05, DLTP08}. 
%We also refer the reader to \cite{SBPP06, Ar13} and the references therein . 

% several document generative models 
%Ralashionship with latent smantic indexing \dots ? 

\subsection{Hyperspectral Unmixing -- Identify Endmembers and Classify Pixels} \label{hsi}

%A hyperspectral image is an image 

Let the columns of the nonnegative data matrix $X$ be the spectral signatures of the pixels in a scene being imaged. 
The spectral signature of a pixel is the fraction of incident light being reflected by that pixel at different wavelengths, and is therefore nonnegative. 
For a hyperspectral image,  there are usually between 100 and 200 wavelength-indexed bands, observed in much broader spectrum than the visible light. %there are usually between 100 and 200 wavelengths, much more than the usual three visible bands in color images. 
This allows for more accurate analysis of the scene under study.  

\begin{figure}[ht!]
\begin{center}
\[
\hspace{-0.5cm} 
\underbrace{X(:,j)}_{\text{\begin{tabular}{c} spectral signature \\ of $j$th pixel \vspace{0.5cm}\\ \includegraphics[width=4.5cm]{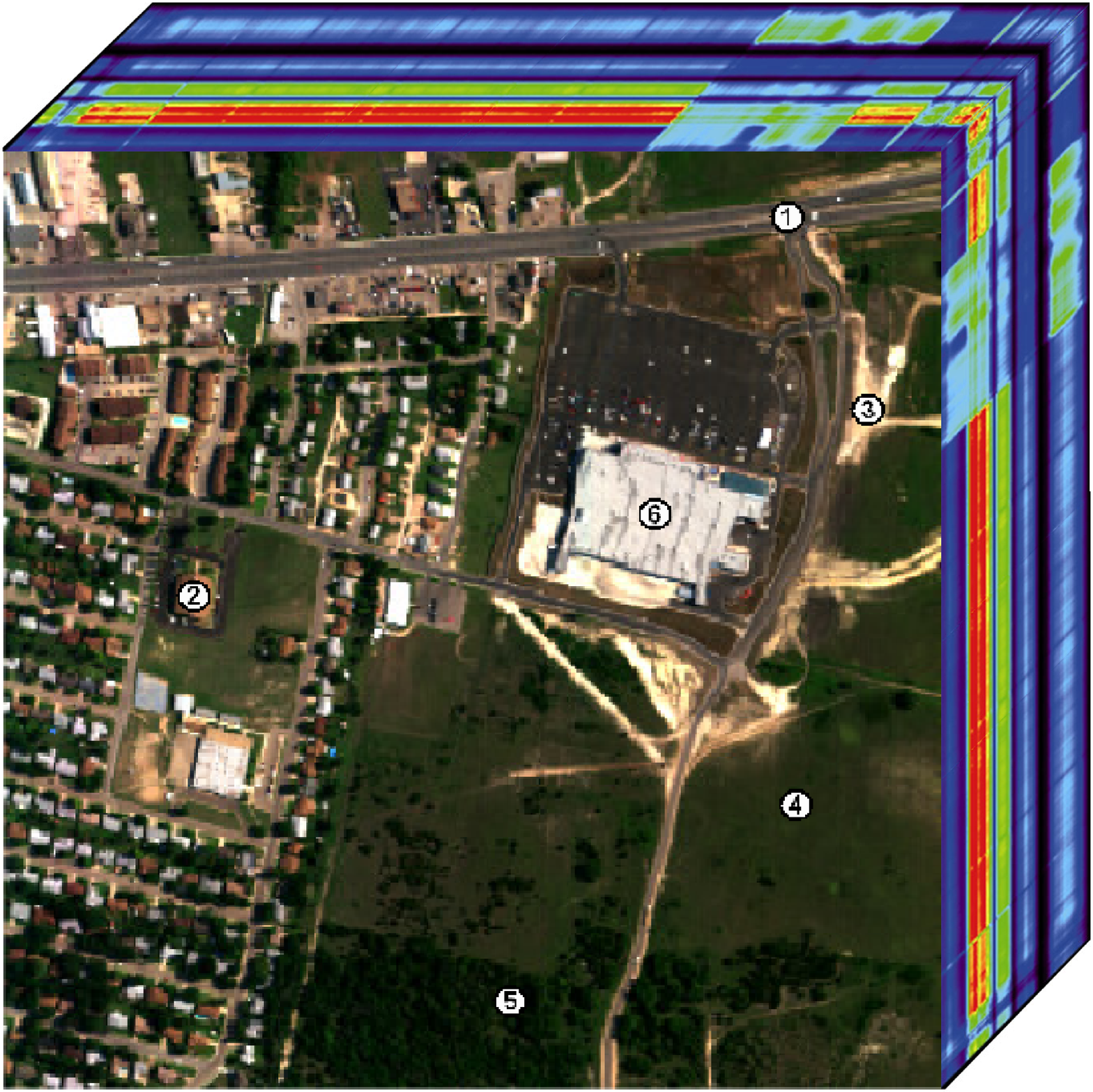} \end{tabular}}} 
\approx   \quad 
\sum_{k=1}^r  
\underbrace{W(:,k)}_{\text{\begin{tabular}{c} spectral signature \\ of $k$th endmember \vspace{0.2cm} \\ \includegraphics[height=6.5cm]{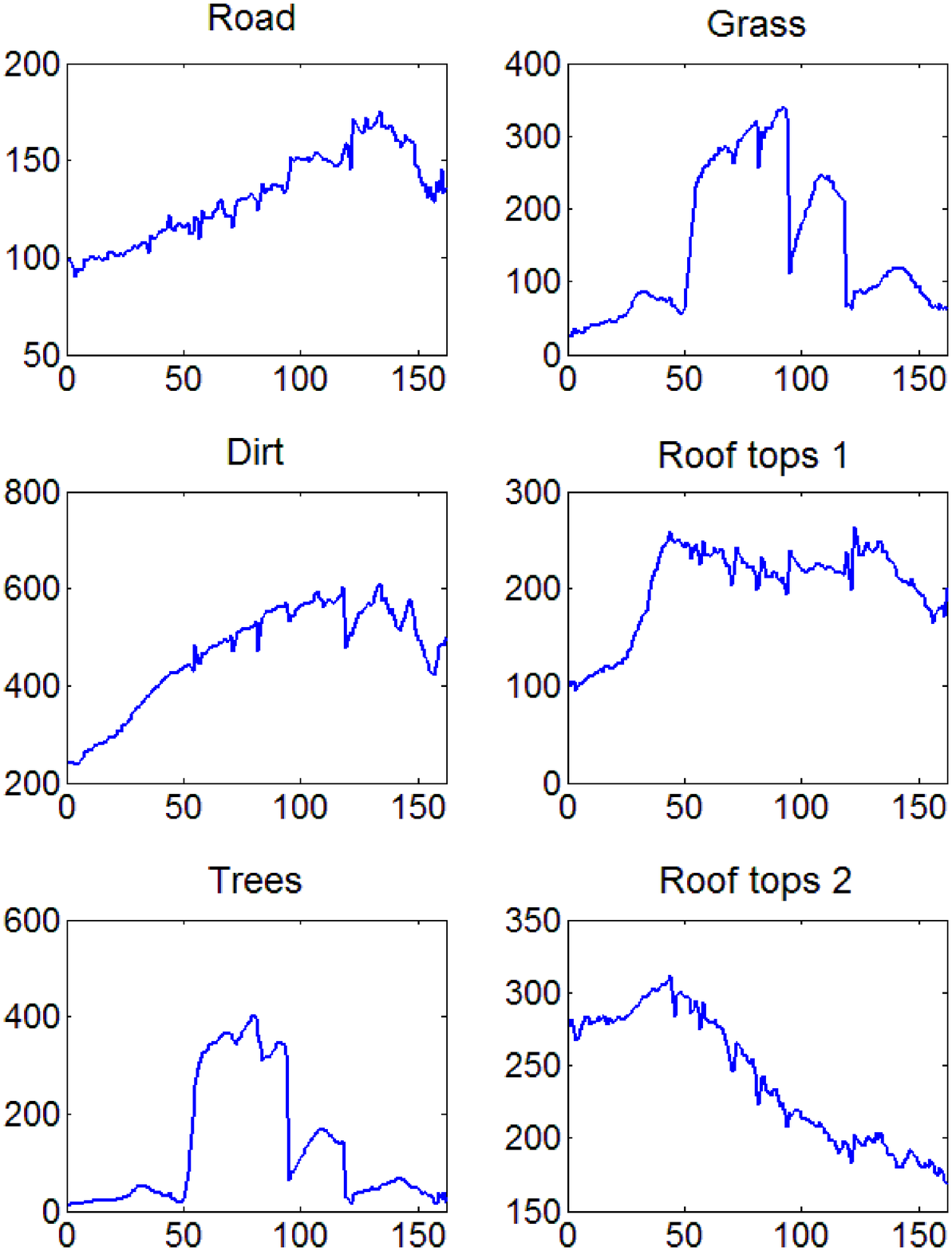} \end{tabular}}}   
\underbrace{H(k,j)}_{\text{\begin{tabular}{c} abundance of $k$th endmember \\ in $j$th pixel \vspace{0.2cm} \\  \includegraphics[height=6.5cm]{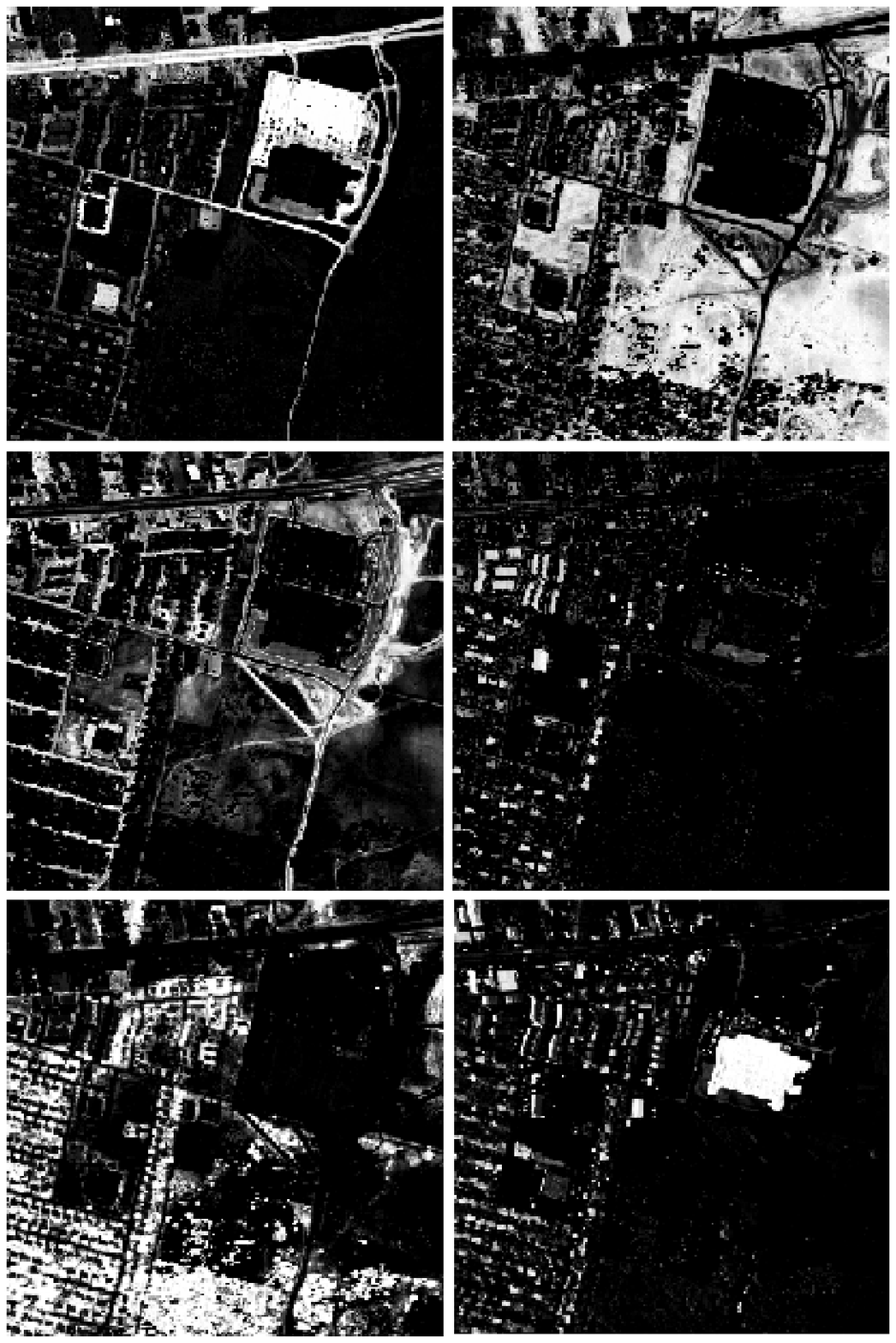} \end{tabular}}} \hspace{-1cm}.  
\]
\caption{\small Decomposition of the Urban hyperspectral image from \protect\url{http://www.agc.army.mil/}, constituted mainly of six endmembers ($r = 6$). Each column of the matrix $W$ is the spectral signature of an endmember, while each row of the matrix $H$ is the abundance map of the corresponding endmember, that is, it contains the abundance of all pixels for that endmember. 
(Note that to obtain this decomposition, we used a sparse prior on the matrix $H$; see Section~\ref{how}.)} 
\label{urban}
\end{center}
\end{figure}

Given a hyperspectral image (see Figure~\ref{urban} for an illustration), the goal of blind hyperspectral unmixing (blind HU) is two-fold: 
\begin{enumerate} 

\item Identify the constitutive materials present in the image; for example, it could be grass, roads, or metallic surfaces. 
These are referred to as the \emph{endmembers}. 

\item Classify the pixels, that is, identify which pixel contains which endmember and in which proportion. (In fact, pixels are in general mixture of several endmembers, due for example to low spatial resolution or mixed materials.)  

\end{enumerate}  
The simplest and most popular model used to address this problem is the \emph{linear mixing model}. It assumes that 
the spectral signature of a pixel results from the linear combination of the spectral signature of the endmembers it contains. 
The weights in the linear combination correspond to the abundances of these endmembers in that pixel. 
For example, if a pixel contains 30\% of grass and 70\% of road surface, then, under the linear mixing model, its spectral signature will be 0.3 times the spectral signature of the grass plus 0.7 times the spectral signature of the road surface. 
This is exactly the NMF model: the spectral signatures of the endmembers are the basis elements, that is, the columns of $W$, while the abundances of the endmembers in each pixel are the weights, that is, the columns of $H$. Note that the factorization rank $r$ corresponds to the number of endmembers in the hyperspectral image. Figure~\ref{urban} illustrates such a decomposition. 
 \\ 

Therefore, given a hyperspectral image, \emph{NMF is able to compute the spectral signatures of the endmembers and simultaneously the abundance of each endmember in each pixel}. 
% which pixels contain which endmember and in which quantity}. 
We refer the reader to~\cite{BP12, Ma14} for recent surveys on blind HU techniques.

\section{The How -- Some Algorithms} \label{how}

We have seen in the previous section that NMF is  a useful LDR technique for nonnegative data. 
The question is now: can we compute such factorizations? 
In this paper, we focus on the following optimization problem 
\begin{equation} \label{nmf}
\min_{W \in \mathbb{R}^{p \times r}, H \in \mathbb{R}^{r \times n}} || X - WH ||_F^2 
\quad 
\text{ such that } 
\quad 
W \geq 0 \text{ and } H \geq 0. 
\end{equation}
Hence we implicitly assume Gaussian noise on the data; see Introduction.  
Although this NMF model is arguably the most popular, it is not always reasonable to assume Gaussian noise for nonnegative data, 
especially for sparse matrices such as document data sets; see the discussion in~\cite{CK12}. 
In fact, many other objective functions are used in practice, e.g., 
the (generalized) Kullback-Leibler divergence for text mining \cite{CK12}, 
the Itakura-Saito distance for music analysis \cite{FCB09}, 
the $\ell_1$ norm to improve robustness against outliers \cite{KK05}, and
the earth mover's distance for computer vision tasks \cite{SL09}. 
Other NMF models are motivated by statistical considerations; we refer the reader to the recent survey~\cite{SFMM14}.

%. The factors are modeled by some joint statistical distribution, then their hyperparameters are modeled by another distributions, usually conjugate priors, etc. In the statistical framework, the posteriors are estimated using well-known tools for statistical inference, e.g. Expectations-Maximization algorithm, Gibbs sampler or Variational Bayes. There are many papers on these topics. See, e.g. by names (A. T. Cemgil, C. Fevotte, A. Ozerov, A. Leijon). Maybe, you can mention about this approach in one short paragraph.  

There are many issues when using NMF in practice. In particular, 
\begin{itemize}

\item \textbf{NMF is NP-hard}. 
Unfortunately, as opposed to the unconstrained problem which can be solved efficiently using the SVD, NMF is NP-hard in general~\cite{V09}. Hence, in practice, most algorithms are applications of standard nonlinear optimization methods and may only be guaranteed to converge to stationary points; see Section~\ref{algo}.  
However, these heuristics have been proved to be successful in many applications. 
More recently, Arora et al.~\cite{AGKM11} described a subclass of nonnegative matrices for which NMF can be solved efficiently. These are the near-separable matrices which will be addressed in Section~\ref{nsnmf}. 
%*** Notes on complexity results? 
Note that Arora et al.~\cite{AGKM11} also described an algorithmic approach for exact NMF\footnote{Exact NMF refers to the NMF problem where an exact factorization is sought: $X = WH$ with $W \geq 0$ and $H \geq 0$.} requiring $\mathcal{O}\big( (pn)^{2^r r^2} \big)$ operations --later improved to $\mathcal{O}\big( (pn)^{r^2} \big)$ by Moitra \cite{Moit13}-- 
hence polynomial in the dimensions $p$ and $n$ for $r$ fixed. 
Although $r$ is usually small in practice, this approach cannot be used to solving real-world problems because of its high computational cost 
(in contrast, most heuristic NMF algorithms run in $\mathcal{O}(pnr)$ operations; see Section~\ref{algo}).

\item \textbf{NMF is ill-posed}. Given an NMF $(W,H)$ of $X$, there usually exist equivalent NMF's $(W',H')$ with $W'H' = WH$. In particular, any matrix $Q$ satisfying $WQ \geq 0$ and $Q^{-1}H \geq 0$ generates such an equivalent factorization. 
The matrix $Q$ can always be chosen as the permutation of a diagonal matrix with positive diagonal elements (that is, as a monomial matrix) and this amounts to the scaling and permutation of the rank-one factors $W(:,k)H(k,:)$ for $1 \leq k \leq r$; this is not an issue in practice.  The issue is when there exist non-monomial matrices $Q$ satisfying the above conditions. In that case, such equivalent factorizations generate different interpretations: for example, in text mining, they would lead to different topics and classifications; 
see the discussion in~\cite{G12}. Here is a simple example 
\[
\left( \begin{array}{cccc}
0 & 1 & 1 & 1 \\
1 & 0 & 1 & 1 \\
1 & 1 & 0 & 1 \\
\end{array} \right) = 
 \left( \begin{array}{ccc}
0 & 1 & 1  \\
1 & 0 & 1  \\
1 & 1 & 0  \\
\end{array} \right)
\left( \begin{array}{cccc}
1 & 0 & 0 & 0.5 \\
0 & 1 & 0 & 0.5\\
0 & 0 & 1 & 0.5 \\
\end{array} \right)
= 
\left( \begin{array}{ccc}
1 & 0 & 0  \\
0 & 1 & 0 \\
0 & 0 & 1  \\
\end{array} \right)
\left( \begin{array}{cccc}
0 & 1 & 1 & 1 \\
1 & 0 & 1 & 1 \\
1 & 1 & 0 & 1 \\
\end{array} \right). 
\]
We refer the reader to \cite{HSS14} and the references therein for recent results on non-uniqueness of NMF.

In practice, this issue is tackled using other priors on the factors $W$ and $H$ and adding proper regularization terms in the objective function. The most popular prior is sparsity which can be tackled with projections~\cite{Hoy} 
or with $\ell_1$-norm penalty~\cite{KP07, G12}. 
For example, in blind HU (Section~\ref{hsi}), 
the abundance maps (that is, the rows of matrix $H$) are usually very sparse (most pixels contain only a few endmembers) and applying plain NMF \eqref{nmf} usually gives poor results for these data sets. 
Other priors for blind HU include piece-wise smoothness of the spectral signatures or spatial coherence (neighboring pixels are more likely to contain the same materials) which are usually tackled with TV-like regularizations (that is, $\ell_1$ norm of the difference between neighboring values to preserve the edges in the image); see, e.g.,~\cite{JQ09, IBP12}, and the references therein. 
Note that the design and algorithmic implementation of refined NMF models for various applications is a very active area of research, e.g., 
graph regularized NMF~\cite{CH11}, 
orthogonal NMF~\cite{Ch08}, 
tri-NMF~\cite{DH06, LZS09}, 
semi and convex NMF~\cite{DJ10}, 
projective NMF \cite{YO10}, 
minimum volume NMF \cite{MH07}, 
and hierarchical NMF~\cite{NBM13}, to cite only a few.

\item \textbf{Choice of $r$.} The choice of the factorization rank $r$, that is, the problem of order model selection, 
is usually rather tricky.  
Several popular approaches are: 
trial and error (that is, try different values of $r$ and pick the one performing best for the application at hand), 
estimation using the SVD (that is, look at the decay of the singular values of the input data matrix), and the use of experts insights (e.g., in blind HU, experts might have a good guess for the number of endmembers present in a scene); see 
also~\cite{BN05, TF09, KS10} and the references therein. 

\end{itemize}

In this section, we focus on the first issue. In Section~\ref{algo}, we present several standard algorithms for the general problem \eqref{nmf}. In Section~\ref{nsnmf}, we describe the near-separable NMF problem and several recent algorithms.

\subsection{Standard NMF Algorithms} \label{algo}

Almost all NMF algorithms designed for \eqref{nmf} use a two-block coordinate descent scheme (exact or inexact; see below), that is, they optimize alternatively over one of the two factors, $W$ or $H$, while keeping the other fixed. The reason is that the subproblem in one factor is convex. 
More precisely, it is a nonnegative least squares problem (NNLS): for example, for $H$ fixed, we have to solve 
%\[
$\min_{W \geq 0} ||X-WH||_F^2$. 
%\]
Note that this problem has a particular structure as it can be decomposed into $p$ independent NNLS in $r$ variables since 
\begin{equation} \label{sepr}
||X-WH||_F^2  
= \sum_{i=1}^p  ||X_{i:}-W_{i:}H||_2^2
= \sum_{i=1}^p W_{i:} \left(HH^T\right) W_{i:}^T - 2  W_{i:} \left(HX_{i:}^T\right) + ||X_{i:}||_2^2. 
\end{equation} 
Many algorithms exist to solve the NNLS problem, and NMF algorithms based on two-block coordinate descent differ by which NNLS algorithm is used; see also, e.g., the discussion in \cite{KHP13}. 
It is interesting to notice that the problem is symmetric in $W$ and $H$ since $||X - WH ||_F^2 = ||X^T - H^T W^T||_F^2$. Therefore, we can focus on the update of only one factor and, in fact, most NMF algorithms 
%do not update both matrices $W$ and $H$ simultaneously nor 
use the same update for $W$ and $H$, and therefore adhere to the framework described in Algorithm~\ref{nmfalgo}. 
\renewcommand{\thealgorithm}{CD}
\algsetup{indent=2em}
\begin{algorithm}[ht!]
\caption{Two-Block Coordinate Descent -- Framework of Most NMF Algorithms \label{nmfalgo}}
\begin{algorithmic}[1] 
\REQUIRE Input nonnegative matrix $X \in \mathbb{R}^{p \times n}_+$ and factorization rank $r$. 
\ENSURE $(W,H) \geq 0$: A rank-$r$ NMF of $X \approx WH$. 
    \medskip  
\STATE Generate some initial matrices $W^{(0)} \geq 0$ and $H^{(0)} \geq 0$; see Section~\ref{init}. 
\FOR {$t$ = 1, 2, \dots$^\dagger$ }  
 \STATE  $W^{(t)}$ = update$\left(X, H^{(t-1)}, W^{(t-1)}\right)$. 
%\STATE Compute $W^{(t)}$ such that $\left\|X- W^{(t)} H^{(t-1)}\right\|_F \leq \left\|X- W^{(t-1)} H^{(t-1)}\right\|_F$. 
 \STATE  ${H^{(t)}}^T$ = update$\left(X^T,  {W^{(t)}}^T, {H^{(t-1)}}^T\right)$. 
\ENDFOR 

$^\dagger$See Section~\ref{stop} for stopping criteria.
\end{algorithmic}  
\end{algorithm}  

The update in steps 3 and 4 of Algorithm~\ref{nmfalgo} usually guarantees the objective function to decrease. 
In this section, we describe the most widely used updates, that is, we describe several standard and widely used NMF algorithms, and compare them in Section~\ref{compa}.  
But first we address an important tool to designing NMF algorithms: the optimality conditions.  To simplify notations, we will drop the iteration index $t$. 

%In Section~\ref{algo}, we describe several standard updates and compare them in Section~\ref{compa}; but first we address an important tool to designing NMF algorithms: the optimality conditions.  

\subsubsection{First-Order Optimality Conditions} \label{kkt} 

Given $X$, let us denote $F(W,H) = \frac{1}{2}||X-WH||_F^2$. The first-order optimality conditions for~\eqref{nmf} are
\begin{eqnarray}  
W \geq 0, & \quad \nabla_W F = WHH^T - XH^T \geq 0, & \quad W \circ \nabla_W F = 0, \label{kktc} \\
 H \geq 0, & \quad \nabla_H F = W^TWH - W^TX \geq 0, & \quad H \circ \nabla_H F = 0, \nonumber 
\end{eqnarray}
where $\circ$ is the component-wise product of two matrices. 
Any $(W,H)$ satisfying these conditions is a stationary point of~\eqref{nmf}. 

It is interesting to observe that these conditions give a more formal explanation of why NMF naturally generates sparse solutions~\cite{GG09}: in fact, any stationary point of~\eqref{nmf} is expected to have zero entries because of the conditions $W \circ \nabla_W F = 0$ and $H \circ \nabla_H F = 0$, that is, the conditions that for all $i,k$ either $W_{ik}$  is equal to zero or the partial derivative of $F$ with respect to $W_{ik}$ is, and similarly for $H$.

\subsubsection{Multiplicative Updates} \label{mu}

Given $X$, $W$ and $H$, the multiplicative updates (MU) modify $W$ as follows 
\begin{equation} %\tag{MU} 
\label{mup}
W \leftarrow W \circ \frac{\left[XH^T\right]}{\left[WHH^T\right]}
\end{equation}
where %$\circ$ denotes the component-wise product, and 
$\frac{[\,]}{[\,]}$ denotes the component-wise division between two matrices. The MU were first developed in~\cite{DM86} for solving NNLS problems, and later rediscovered and used for NMF in~\cite{LS01}. 
The MU are based on the majorization-minimization framework. In fact, \eqref{mup} is the global minimizer of a quadratic function majorizing $F$, that is, a function that is larger than $F$ everywhere and is equal to $F$ at the current iterate \cite{DM86, LS01}. Hence minimizing that function guarantees $F$ to decrease and therefore leads to an algorithm for which $F$ monotonically decreases. 
The MU can also be interpreted as a rescaled gradient method: in fact, 
\begin{equation} \label{mugrad}
W \circ \frac{\left[XH^T\right]}{\left[WHH^T\right]} = W -  \frac{\left[W\right]}{\left[WHH^T\right]} \circ  \nabla_W F . 
\end{equation}
Another more intuitive interpretation is as follows: we have that 
\[
\frac{\left[XH^T\right]_{ik}}{\left[WHH^T\right]_{ik}} \geq 1 \quad \iff \quad \left(\nabla_{W} F\right)_{ik} \leq 0.
\] 
Therefore, in order to satisfy \eqref{kktc}, for each entry of $W$, the MU either 
%\begin{itemize}
%\item 
(i)~increase it if its partial derivative is negative,   
%\item 
(ii)~decrease it if its partial derivative is positive,  or 
%\item 
(iii)~leave it unchanged if its partial derivative is equal to zero. 
%\end{itemize} 

If an entry of $W$ is equal to zero, the MU cannot modify it hence it may occur that an entry of $W$ is equal to zero while its partial derivative is negative which would not satisfy \eqref{kktc}.  
Therefore, the MU are not guaranteed to converge to a stationary point\footnote{If the initial matrices are chosen positive, 
some entries can first converge to zero while their partial derivative eventually becomes negative or zero (when strict complementarity is not met) which is numerically unstable; see \cite{GG12} for some numerical experiments.}. 
There are several ways to fix this issue, e.g., 
rewriting the MU as a rescaled gradient descent method --see Equation~\eqref{mugrad}: only entries in the same row interact-- and modifying the step length~\cite{Lin07}, or 
using a small positive lower bound for the entries of $W$ and $H$~\cite{GG12, TR13}; see also~\cite{BBV}. 
A simpler and nice way to guarantee convergence of the MU to a stationary point is proposed in~\cite{CK12}: 
use the original updates \eqref{mup} while reinitializing zero entries of $W$ to a small positive constant when their partial derivatives become negative. 

  The MU became extremely popular mainly because 
	(i)~they are simple to implement\footnote{For example, in Matlab: \texttt{W = W.*(X*H')./(W*(H*H'))}. \label{fn2}}, 
	(ii)~they scale well and are applicable to sparse matrices\footnote{When computing the denominator $WHH^T$ in the MU, it is crucial to compute $HH^T$ first in order to have the lowest computational cost, and make the MU scalable for sparse matrices; see, e.g., footnote~\ref{fn2}.}, 
	and 
	(iii)~they were proposed in the paper of Lee and Seung~\cite{LS99} which launched the research on NMF. 
	However, the MU converge relatively slowly; see, e.g.,~\cite{Han09} for a theoretical analysis, and Section~\ref{compa} for some numerical experiments.  
	Note that the original MU only update $W$ once before updating $H$. They can be significantly accelerated using a more effective alternation strategy~\cite{GG12}: the idea is to update $W$ several times before updating $H$ because the products $HH^T$ and $XH^T$ do not need to be recomputed.

\subsubsection{Alternating Least Squares}

The alternating least squares method (ALS) first computes the optimal solution of the unconstrained least squares problem $\min_W ||X-WH||_F$ and then project the solution onto the nonnegative orthant: 
\begin{equation*}  
W \leftarrow \max\Big(  \argmin_{Z \in \mathbb{R}^{p \times r}} ||X-ZH||_F , 0 \Big), 
\end{equation*} 
where the $\max$ is taken component-wise. 
The method has the advantage to be relatively cheap, 
and easy to implement\footnote{For example, in Matlab: \texttt{W = max(0,(X*H')/(H*H'))}. \label{fn1}}. 
ALS usually does not converge: the objective function of \eqref{nmf} might oscillate under the ALS updates (especially for dense input matrices $X$; see Section~\ref{compa}). It is interesting to notice that, because of the projection, the solution generated by ALS is not scaled properly. In fact, the error can be reduced (sometimes drastically) by multiplying the current solution $WH$ by the constant 
\begin{equation} 
\alpha^* 
= \argmin_{\alpha \geq 0} ||X - \alpha WH||_F 
= \frac{\langle X , WH\rangle}{\langle WH , WH\rangle}
= \frac{\langle XH^T , W\rangle}{\langle W^TW , HH^T\rangle}.  \label{scal}
\end{equation}
Although it is in general not recommended to use ALS because of the convergence issues, ALS can be rather powerful for initialization purposes (that is, perform a few steps of ALS and then switch to another NMF algorithm), especially for sparse matrices~\cite{CAZP09}.

\subsubsection{Alternating Nonnegative Least Squares} \label{anls}

Alternating nonnegative least squares (ANLS) is a class of methods where the subproblems in $W$ and $H$ are solved exactly, that is, the update for $W$ is given by  
\begin{equation*} %\tag{ANLS}
W  \leftarrow  \argmin_{W \geq 0} ||X-WH||_F .  
\end{equation*} 
Many methods can be used to solve the NNLS $\argmin_{W \geq 0} ||X-WH||_F$, 
and dedicated active-set methods have shown to perform very well in practice\footnote{In particular, the Matlab function \texttt{lsqnonneg} implements an active-set method from \cite{LH74}.}; see~\cite{KP07, KP08, KP11}. Other methods are based for example on projected gradients~\cite{L07}, Quasi-Newton~\cite{CZA06}, or fast gradient methods~\cite{GTA12}. 
ANLS is guaranteed to converge to a stationary point~\cite{GS00}. 
Since each iteration of ANLS computes an optimal solution of the NNLS subproblem, each iteration of ANLS decreases the error the most among NMF algorithms following the framework described in Algorithm~\ref{nmfalgo}. However, each iteration is computationally more expensive, and more difficult to implement. 

%Note that Because at the first steps of ANLS, it does not make much sens to optimize $W$ exactly, it would make more sens to use it as a

Note that, because usually the initial guess $WH$ is a poor approximation of $X$, it does not make much sense to solve the NNLS subproblems exactly at the first steps of Algorithm~\ref{nmfalgo}, and therefore it might be profitable to use ANLS rather in a refinement step of a cheaper NMF algorithm (such as the MU or ALS).

\subsubsection{Hierarchical Alternating Least Squares} \label{hals}

Hierarchical alternating least squares (HALS) solves the NNLS subproblem using an exact coordinate descent method, updating one column of $W$ at a time. The optimal solutions of the corresponding subproblems can be written in closed form. In fact, the entries of a column of $W$ do not interact --see Equation~\eqref{sepr}-- hence the corresponding problem can be decoupled into $p$ quadratic problems with a single nonnegative variable.  HALS updates $W$ as follows. 
For $\ell = 1, 2, \dots, r$: 
\begin{align*}
W(:,\ell) 
& \leftarrow 
\argmin_{W(:,\ell) \geq 0} \Big\|X-\sum_{k\neq \ell} W(:,k)H(k,:) - W(:,\ell)H(\ell,:) \Big\|_F  \\ 
& \leftarrow \max \left(0,  \frac{XH(\ell,:)^T -\sum_{k\neq \ell} W(:,k) \left(H(k,:) H(\ell,:)^T\right)   }{ ||H(\ell,:)||_2^2 } \right) .  %\tag{HALS} 
\end{align*} 
%HALS therefore updates $W$ using a block-coordinate descent method with $r$ blocks of $m$ variables. 
HALS has been rediscovered several times, 
originally in~\cite{CZA07} (see also~\cite{CP09b}), 
then as the rank-one residue iteration (RRI)~in~\cite{Ho08}, 
as FastNMF in~\cite{LZ09}, and also in~\cite{LW12}. 
Actually, HALS was first described in Rasmus Bro's thesis \cite[pp.161-170]{Bro98} (although it was not investigated thoroughly): 
\begin{quote} 
\dots to solve for a column vector $w$ of $W$ it is only necessary to solve the unconstrained problem and subsequently set negative values to zero. 
Though the algorithm for imposing non-negativity is thus simple and may be advantageous in some situations, it is not pursued here. Since it optimizes a smaller subset of parameters than the other approaches it may be unstable in difficult situations. 
\end{quote}
 HALS was observed to converge much faster than the MU (see \cite[p.131]{NG11} for a theoretical explanation, and Section~\ref{compa} for a comparison) while having almost the same computational cost; see~\cite{GG12} for a detailed account of the flops needed per iteration. 
%To the best of our knowledge, HALS is among the best NMF algorithms, and combines cheap updates with a relatively high decrease in the objective function at each step. 
Moreover, HALS is, under some mild assumptions, guaranteed to converge to a stationary point; see the discussion in~\cite{GG12}. 
Note that one should be particularly careful when initializing HALS otherwise the algorithm could set some columns of $W$ to zero initially (e.g., if $WH$ is badly scaled with $WH \gg X$) hence it is recommended to initially scale $(W,H)$ according to \eqref{scal}; see the discussion in \cite[p.72]{NG11}.  

%An explanation of the good behavior of HALS is as follows: the conditioning of NNLS good because $HH^T$ is usually

In the original HALS, each column of $W$ is updated only once before updating $H$. 
However, as for the MU, it can be sped up by updating $W$ several times before updating $H$~\cite{GG12}, or selecting the entries of $W$ to update following a Gauss-Southwell-type rule~\cite{HD11}. 
HALS can also be generalized to other cost functions using Taylor expansion~\cite{LLP12}.

\subsubsection{Comparison} \label{compa}

\begin{figure}[ht!]
\begin{center}
\begin{tabular}{cc}
\includegraphics[width=8cm]{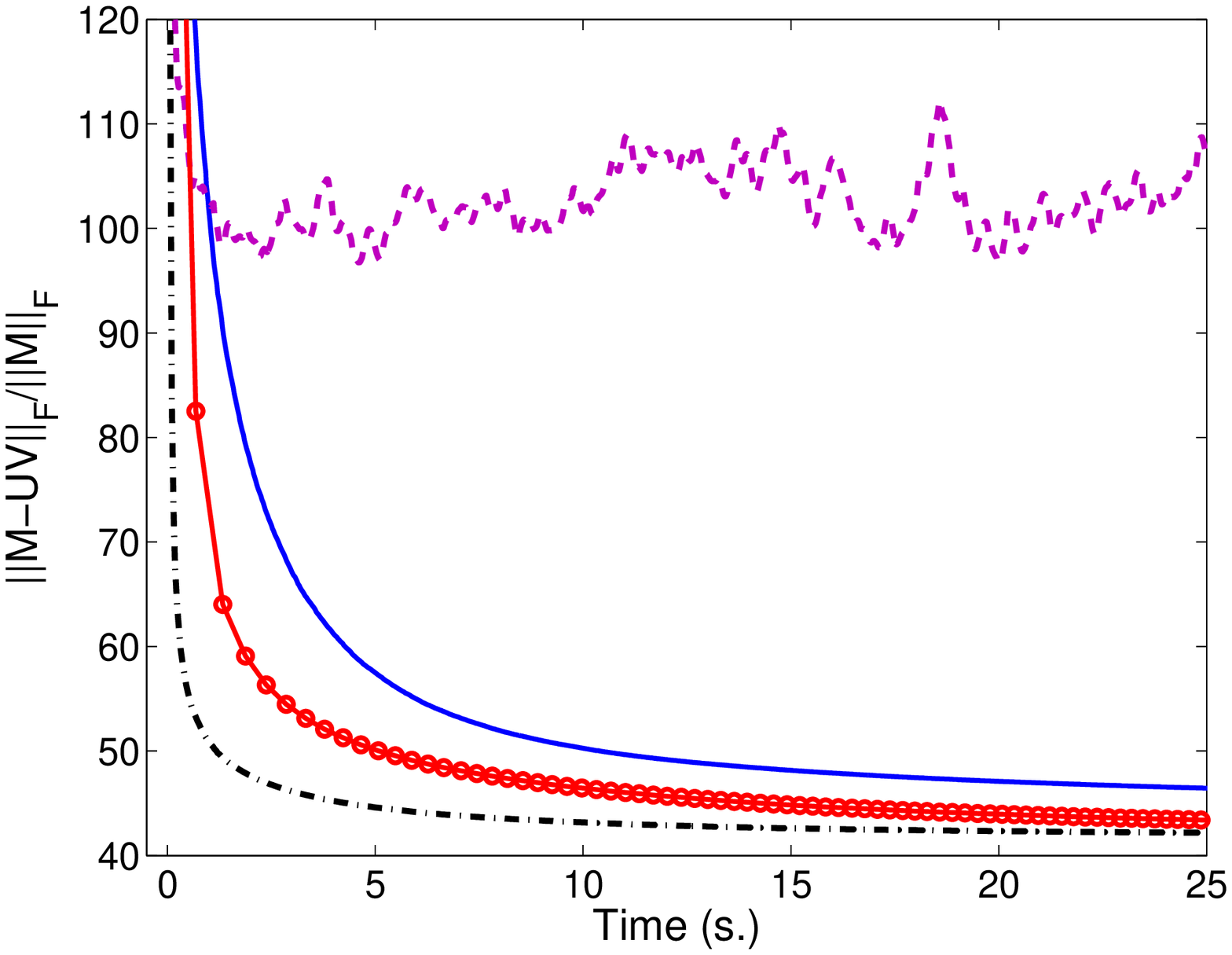}  & \includegraphics[width=8cm]{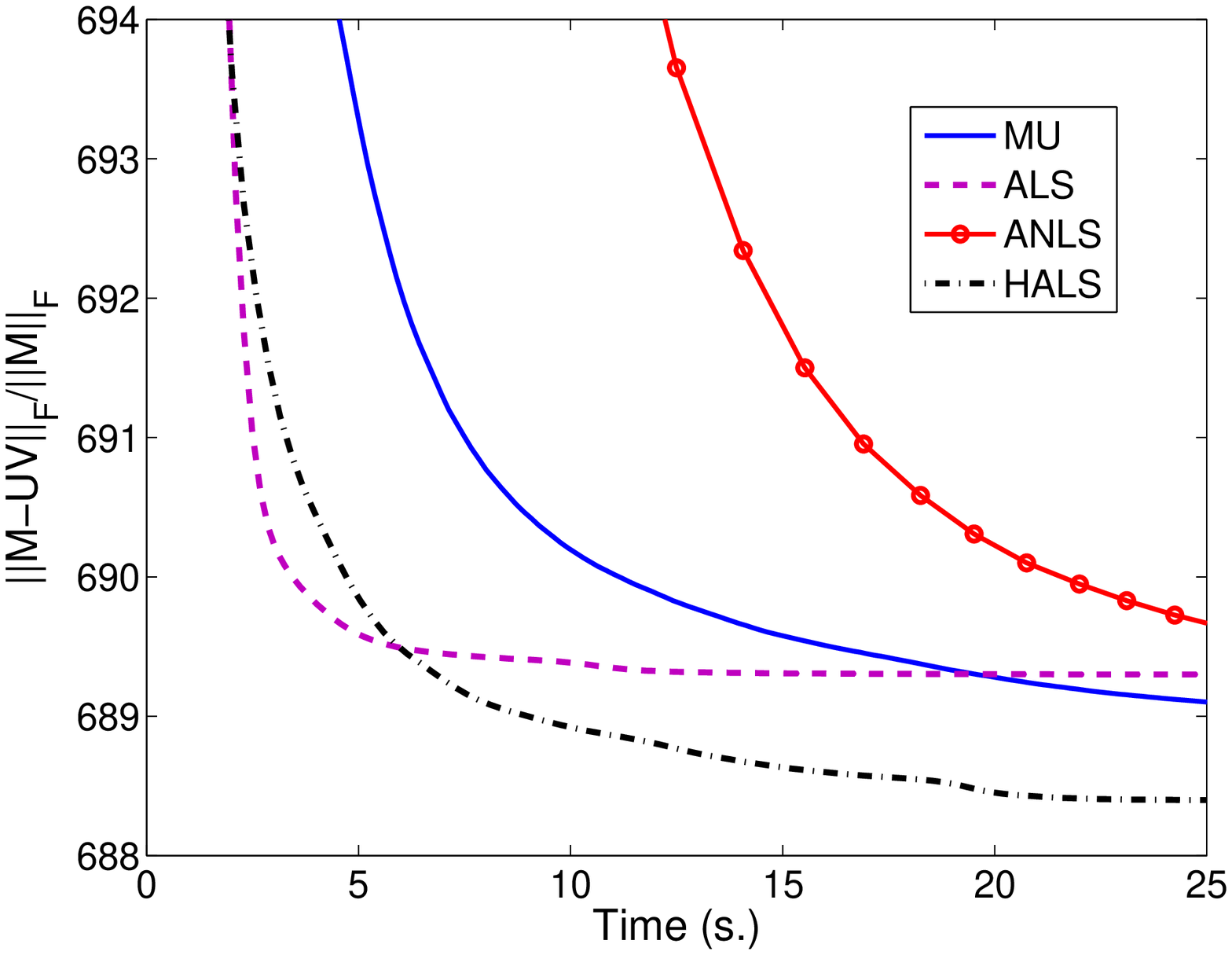}  \\
\end{tabular}
\end{center} 
\caption{\small Comparison of MU, ALS, ANLS and HALS. On the left: CBCL facial images with $r=49$; same data set as in Figure~\ref{cbcl}. 
On the right: Classic document data set with $m=7094$, $n = 41681$ and $r= 20$; see, e.g.,~\cite{ZG05}.  
The figure displays the average error using the same ten initial matrices $W$ and $H$ for all algorithms, randomly generated with the \texttt{rand} function of Matlab. All tests were performed using Matlab on a laptop Intel CORE i5-3210M CPU @2.5GHz 2.5GHz 6Go RAM. Note that, for ALS, we display the error after scaling; see Equation~\eqref{scal}. For MU and HALS, we used the implementation from \protect\url{https://sites.google.com/site/nicolasgillis/}, for ANLS from \protect\url{http://www.cc.gatech.edu/~hpark/nmfsoftware.php}, and ALS was implemented following footnote~\ref{fn1}.} 
\label{algocomp}
\end{figure}

Figure~\ref{algocomp} displays the evolution of the objective function of \eqref{nmf} for the algorithms described in the previous section: 
on the left,  
the dense CBCL data set (see also Figure~\ref{cbcl}), 
and, 
on the right, 
the sparse Classic document data set.   
As anticipated in the description of the different algorithms in the previous sections,  we observe that: 
\begin{itemize}
\item The MU converge rather slowly. 
\item ALS oscillates for the dense matrix (CBCL data set)  and performs quite poorly while, for the sparse matrix (Classic data set), 
it converges initially very fast but then stabilizes and cannot compute a solution with small objective function value.  
\item ANLS performs rather well for the dense matrix and is the second best after HALS. However, it performs rather poorly for the sparse matrix. 
\item HALS performs the best as it generates the best solutions within the allotted time. 
\end{itemize}
For other comparisons of NMF algorithms and more numerical experiments, we refer the reader to the book~\cite{CAZP09}, the theses 
\cite{Ho08, NG11}, the survey~\cite{BBLPP07}, and the references therein. \\ 
%Moreover, as noted aove, simple hybridation strategies could improve performance; for example using ALS + ANLS on sparse matrices. \\ 

Further research on NMF includes the design of more efficient algorithms, in particular for regularized problems; 
see, e.g., \cite{RBL13} for a recent example of imposing sparsity in a more robust and stable way. 
We conclude this section with some comments about stopping criteria and initializations of NMF algorithms.

\subsubsection{Stopping Criterion} \label{stop}

There are several approaches for the stopping criterion of NMF algorithms, as in usual non-linear optimization schemes, e.g., 
based on the evolution of the objective function, on the optimality conditions~\eqref{kktc}, 
or on the difference between consecutive iterates. 
These criteria are typically combined with either a maximum number of iterations or a time limit to ensure
termination; see, e.g., the discussion in~\cite{NG11}. 
In this section, we would like to point out an issue which is sometimes overlooked in the literature when using the optimality conditions to assess the convergence of NMF algorithms. 
A criterion based on the optimality conditions is for example $C(W,H) = C_W(W) + C_H(H)$ where 
\begin{equation} \label{stopcriteq}
C_W(W) = \underbrace{||\min(W,0)||_F}_{W \geq 0} 
+ 
\underbrace{||\min(\nabla_W F,0)||_F}_{\nabla_W F \geq 0} 
+ 
\underbrace{||W \circ \nabla_W F||_F}_{W \circ \nabla_W F = 0}  , 
\end{equation}
and similarly $C_H(H)$ for $H$, so that $C(W,H) = 0 \iff (W,H)$ is a stationary point of \eqref{nmf}.  
There are several problems to using $C(W,H)$ (and other similar variants) as a stopping criterion  and for comparing the convergence of different algorithms: 

\begin{itemize}

\item It is sensitive to scaling. For $\alpha > 0$ and $\alpha \neq 1$, we will have in general that 
\[
C_W(W) + C_H(H) = C(W,H) \neq C(\alpha W, \alpha^{-1}H). 
\] 
since the first two terms in~\eqref{stopcriteq} are sensitive to scaling. 
For example, if one solves the subproblem in $W$ exactly and obtains $C_W(W) = 0$ (this will be the case for ANLS; see Section~\ref{anls}), then $\nabla_H F$ can be made arbitrarily small by multiplying $W$ by a small constant and dividing $H$ by the same constant (while, if $H \geq 0$, it will not influence the first term which is equal to zero). 
This issue can be handled with proper normalization, e.g., imposing $||W(:,k)||_2 = ||H(k,:)||_2$ for all $k$; see \cite{Ho08}. 

\item The value of $C(W,H)$ after the update of $W$ can be very different from the value after an update of $H$ (in particular, 
if the scaling is bad or if $|m - n| \gg 0$). Therefore, one should be very careful when using this type of criterion to compare ANLS-type methods with other algorithms such as the MU or HALS as the evolution of $C(W,H)$ can be misleading (in fact, an algorithm that monotonically decreases the objective function, such as the MU or HALS, is not guaranteed to monotonically decrease $C(W,H)$.) A potential fix would be to scale the columns of $W$ and the rows of $H$ so that $C_W(W)$ after the update of $H$ and $C_H(H)$ after the update of $W$ have the same order of magnitude. 
%Another possibility would be to apply one step of ANLS before evaluating the criterion (hence implicitly focusing on one of the two factors since after an update of ANLS, one of the two terms, $C_W(W)$ or $C_H(H)$, is equal to zero). 

\end{itemize} 

%However, it is not clear at this point whether there exists a good criterion based on the optimality conditions for comparing convergence of different NMF algorithms. 

\subsubsection{Initialization}  \label{init}

A simple way to initialize $W$ and $H$ is to generate them randomly (e.g., generating all entries uniformly at random in the interval [0,1]). 
Several more sophisticated initialization \index{initialization} strategies have been developed in order to have better initial estimates in the hope to 
(i)~obtain a good factorization with fewer iterations, and 
(ii)~converge to a better stationary point. However, most initialization strategies come with no theoretical guarantee 
(e.g., a bound on the distance of the initial point to optimality) which can be explained in part by the complexity of the problem (in fact, NMF is NP-hard in general --see the introduction of this section). This could be an interesting direction for further research. 
We list some initialization strategies here, they are based on  
\begin{itemize}

\item \emph{Clustering techniques}. 
Use the centroids computed with some clustering method, e.g., with $k$-means or spherical $k$-means, to initialize the columns of $W$, and initialize $H$ as a proper scaling of the cluster indicator matrix (that is, $H_{kj} \neq 0 \iff X(:,j)$ belongs to the $k$th cluster)~\cite{Wild04, X08}; see also~\cite{Casa13} and the references therein for some recent results. 

\item \emph{The SVD}. 
Let $\sum_{k=1}^r  u_k v_k^T$ be the best rank-$r$ approximation  of $X$ (which can be computed, e.g., using the SVD; see Introduction). 
Each rank-one factor $u_k v_k^T$ might contain positive and negative entries (except for the first one, by the Perron-Frobenius theorem\footnote{Actually, the first factor could contain negative entries if the input matrix is reducible and its first two singular values are equal to one another; see, e.g., \cite[p.16]{NG11}.}). 
However, denoting $[x]_+ = \max(x,0)$, we have 
\[
u_k v_k^T = [u_k]_+ [v_k^T]_+ + [-u_k]_+ [-v_k^T]_+ - [-u_k]_+[v_k^T]_+ - [u_k]_+[-v_k^T]_+ , 
\]
and the first two rank-one factors in this decomposition are nonnegative. 
 Boutsidis et al.~\cite{Bou} proposed to replace each rank-one factor in $\sum_{k=1}^r  u_k v_k^T$ with either $[u_k]_+ [v_k^T]_+$ or $[-u_k]_+ [-v_k^T]_+$, selecting the one with larger norm and scaling it properly. 
 %the nonnegative rank-one factor whose norm is larger (that is, $[u_k]_+ [v_k^T]_+$ or $[-u_k]_+ [-v_k^T]_+$) and scale it properly.    

\item \emph{Column subset selection}. It is possible to initialize the columns of $W$ using data points, that is, 
initialize $W = X(:,\mathcal{K})$ for some set $\mathcal{K}$ with cardinality $r$; see \cite{CM11, Zd12} and Section~\ref{nsnmf}. 

\end{itemize}

In practice, one may use several initializations, and keep the best solution obtained; see, e.g., the discussion in \cite{CAZP09}. 

%A potential direction for further research could the design of an algorithm guaranteed to recover the optimal solution (although such an algorithm would have a bad worst-case complexity). 

\subsection{Near-Separable NMF} \label{nsnmf}

A matrix $X$ is $r$-separable if there exists an index set $\mathcal{K}$ of cardinality $r$ such that 
\[
X = X(:,\mathcal{K}) H \qquad \text{ for some } H \geq 0. 
\]
In other words, there exists a subset of $r$ columns of $X$ which generates a convex cone containing all columns of~$X$. 
Hence, given a separable matrix, the goal of separable NMF is to identify the subset of columns $\mathcal{K}$ that allows to reconstruct all columns of $X$ (in fact, given $X(:,\mathcal{K})$, $H$ can be computed by solving a convex optimization program; see Section~\ref{algo}).  
The separability assumption makes sense in several applications: for example, 
\begin{itemize}

\item In text mining (see Section~\ref{text}), separability of the word-by-document matrix requires that for each topic, there exists a document only on that topic. 
Note that we can also assume separability of the transpose of $X$ (that is, of the document-by-word matrix), i.e., for each topic there exists one word used only by that topic (referred to as an `anchor' word). In fact, the latter is considered a more reasonable assumption in practice; 
see~\cite{KSK12, Ar13, DRIS13}  and also the thesis \cite{Ge13} for more details.

\item In hyperspectral unmixing \index{blind hyperspectral unmixing} (see Section~\ref{hsi}),  separability of the wavelength-by-pixel matrix requires that for each endmember there exists a pixel containing only that endmember. 
This is the so-called pure-pixel assumption, and makes sense for relatively high spatial resolution hyperspectral images; see~\cite{BP12, Ma14} and the references therein.  

\end{itemize}
Separability has also been used successfully in 
blind source separation~\cite{NN05, CMCW08}, 
video summarization and image classification~\cite{ESV12},
and foreground-background separation in computer vision \cite{KS13}. 
Note that for facial feature extraction described in Section~\ref{image}, 
separability does not make sense since we cannot expect features to be present in the data set.  \\

It is important to points out that separable NMF is closely related to several problems, including 
\begin{itemize}
\item Column subset selection \index{column subset selection} which is a long-standing problem in numerical linear algebra (see~\cite{BM09} and the references therein). 

\item Pure-pixel search in hyperspectral unmixing which has been addressed long before NMF was introduced; 
see \cite{Ma14} for a historical note. 

\item The problem of identifying a few important data points in a data set (see~\cite{ESV12} and the references therein). 

\item Convex NMF~\cite{DJ10}, and the CUR decomposition~\cite{MD09}. 

%Moreover, as mentioned above, near-separable NMF algorithms can be used as efficient initialization strategies for standard NMF algorithms; see, e.g.,~. 
\end{itemize}
Therefore, it is difficult to pinpoint the roots of separable NMF and a comprehensive overview of all methods related to separable NMF is out of the scope of this paper. 
However, to the best of our knowledge, it is only very recently that provably efficient algorithms for separable NMF have been proposed. 
This new direction of research was launched by a paper by Arora et al.~\cite{AGKM11} which shows that NMF of separable matrices can be computed efficiently (that is, in polynomial time), even in the presence of noise (the error can be bounded in terms of the noise level; see below). We focus in this section on these provably efficient algorithms for separable NMF.  \\

In the noiseless case, separable NMF reduces to identifying the extreme rays of the cone spanned by the columns of $X$.  
If the columns of the input matrix $X$ are normalized so that their entries sum to one, that is, $X(:,j) \leftarrow ||X(:,j)||_1^{-1} X(:,j)$ for all $j$ (and discarding the zero columns of $X$), ~
then the problem reduces to identifying the vertices of the convex hull of the columns of $X$. 
In fact, since the entries of each column of $X$ sum to one and \mbox{$X = X(:,\mathcal{K}) H$}, the entries of each column of $H$ must also sum to one: as $X$ and $H$ are nonnegative, we have for all $j$ 
\begin{align*}
1 = ||X(:,j)||_1 & = ||X(:,\mathcal{K}) H(:,j)||_1 \\
& = \sum_k ||X(:,\mathcal{K}(k))||_1 H(k,j) = \sum_k H(k,j) = ||H(:,j)||_1.  
\end{align*} 
Therefore, the columns of $X$ are convex combinations (that is, linear combinations with nonnegative weights summing to one) of the columns of $X(:,\mathcal{K})$. 

In the presence of noise, the problem is referred to as near-separable NMF\index{near-separable NMF}, and can be formulated as follows 
 \begin{quote}
(Near-Separable NMF) \emph{Given a noisy $r$-separable matrix $\tilde{X} = X + N$ with $X = W[I_r, H'] \Pi$ where $W$ and $H'$ are nonnegative matrices, $\Pi$ is a permutation matrix and $N$ is the noise, find a set $\mathcal{K}$ of $r$ indices such that $\tilde{X}(:,\mathcal{K}) \approx W$. } 
\end{quote} 
In the following, we describe some algorithms for near-separable NMF; they are classified in two categories:  algorithms based on self-dictionary and sparse regression (Section~\ref{sdsr}) and geometric algorithms (Section~\ref{geo}).

\subsubsection{Self-Dictionary and Sparse Regression Framework} \label{sdsr}

In the noiseless case, separable NMF can be formulated as follows 
\begin{equation} \label{spreg}
\min_{Y \in \mathbb{R}^{n \times n}} ||Y||_{\text{row},0}  \quad \text{ such that } \quad X = XY \text{ and } Y \geq 0,  
\end{equation}
where $||Y||_{\text{row},0}$ is the number of non-zero rows of $Y$. 
In fact, 
if all the entries of a row of $Y$ are equal to zero, 
then the corresponding column of $X$ is not needed to reconstruct the other columns of $X$. 
Therefore, minimizing the number of rows of $Y$ different from zero is equivalent to minimizing the number of columns of $X$ used to reconstruct all the other columns of $X$, which solves the separable NMF problem. In particular, given an optimal solution $Y^*$ of \eqref{spreg} and denoting 
$\mathcal{K} = \{ i | Y^*(i,:) \neq 0 \}$, we have $X = WY^*(\mathcal{K},:)$ where $W = X(:,\mathcal{K})$.

In the presence of noise, the constraints $X = XY$ are usually reformulated as $||X-XY|| \leq \delta$ for some $\delta > 0$ or put as a penalty $\lambda ||X-XY||$ in the objective function for some penalty parameter $\lambda > 0$. 
In~\cite{ESV12, EMO12}, $||Y||_{\text{row},0}$ is replaced using $\ell_1$-norm type relaxation: 
\[
||Y||_{q,1} = \sum_j ||Y(i,:)||_q, 
\]
where $q > 1$ so that $||Y||_{q,1}$ is convex and \eqref{spreg} becomes a convex optimization problem. 
Note that this idea is closely related to compressive sensing where $\ell_1$-norm relaxation is used to find the sparsest solution to an underdetermined linear system. This relaxation is exact given that the matrix involved in the linear system satisfies some incoherence properties. In separable NMF, the columns and rows of matrix $X$ are usually highly correlated hence it is not clear how to extend the results from the compressive sensing literature to this separable NMF model; see, e.g., the discussion in \cite{Ma14}.

A potential problem in using convex relaxations of \eqref{spreg} is that it cannot distinguish duplicates of the columns of $W$. In fact, if a column of $W$ is present twice in the data matrix $X$, the corresponding rows of $Y$ can both be non-zero hence both columns of $W$ can potentially be extracted 
(this is because of the convexity and the symmetry of the objective function) --in~\cite{ESV12}, $k$-means is used as a pre-processing in order to remove duplicates. Moreover, although this model was successfully used  to solve real-world problems, 
no robustness results were developed so far so it is not clear how this model behaves in the presence of noise (only asymptotic results were proved, that is, when the noise level goes to zero and when no duplicates are present~\cite{ESV12}).  

A rather different approach to enforce row sparsity was suggested in~\cite{BRRT12}, and later improved in~\cite{GL13}. 
Row sparsity of $Y$ is enforced by 
(i)~minimizing a weighted sum of the diagonal entries of $Y$ hence enforcing $\diag(Y)$ to be sparse (in fact, this is nothing but a weighted $\ell_1$ norm since $Y$ is nonnegative), and 
(ii)~imposing all entries in a row of $Y$ to be smaller than the corresponding diagonal entry of $Y$ (we assume here that the columns of $X$ are normalized). 
The second condition implies that if $\diag(Y)$ is sparse then $Y$ is row sparse. The corresponding near-separable NMF model is: 
\begin{equation} \label{bit}
\min_{Y \in \mathbb{R}^{n \times n}} p^T \diag(Y) 
\quad \text{ such that } \quad 
||X-XY||_1 \leq \delta \quad \text{ and } \quad  0 \leq Y_{ij}  \leq Y_{ii} \leq 1, 
\end{equation}
for some positive vector $p \in \mathbb{R}^n$ with distinct entries (this breaks the symmetry so that the model can distinguish duplicates).  
This model has been shown to be robust: defining the parameter\footnote{The larger the parameter $\alpha$ is, the less sensitive the data to noise. For example, it can be easily checked that $\epsilon = \max_j ||N(:,j)||_1 < \frac{\alpha}{2}$ is a necessary condition to being able to distinguish the columns of $W$~\cite{G13h}.} $\alpha$ as 
\[
\alpha(W) = \min_{1 \leq j \leq r} \min_{x \in \mathbb{R}^{r-1}_+} ||W(:,j) - W(:,\mathcal{J}_j)x||_1, \quad \text{ where } \mathcal{J}_j = \{1,2,\dots,r\} \backslash \{j\}, 
\] 
and for a near-separable matrix $\tilde{X} = W[I_r, H']\Pi + N$ (see above) with $\epsilon = \max_j ||N(:,j)||_1 \leq \mathcal{O}\left(\frac{\alpha^2(W)}{r} \right)$, 
the model~\eqref{bit} can be used to identify the columns of $W$ 
with $\ell_1$ error proportional to $\mathcal{O}\left( \frac{r \epsilon}{\alpha(W)} \right)$, that is, the identified index set $\mathcal{K}$  satisfies 	$\max_j \min_{k \in \mathcal{K}} ||\tilde{X}(:,k) - W(:,j)||_{1} \leq \mathcal{O}\left(  \frac{r \epsilon}{\alpha(W)} \right)$; see 
\cite[Th.7]{GL13} for more details. \\

Finally, a drawback of the approaches based on self-dictionary and sparse regression is that they are computationally expensive as they require to tackle optimization problems with $n^2$ variables.

\subsubsection{Geometric Algorithms}  \label{geo}

Another class of near-separable algorithms are based on geometric insights and in particular on the fact that the columns of $W$ are the vertices of the convex hull of the normalized columns of $X$.  
The first geometric algorithms can be found in the remote sensing literature (they are referred to as endmember extraction algorithms or pure-pixel identification algorithms), see~\cite{Ma14} for a historical note; 
and~\cite{BP12} for a comprehensive survey. 
Because of the large body of literature, we do not aim at surveying all algorithms but rather focus on a single algorithm which is particularly simple while being rather effective in practice:  the successive projection algorithm (SPA).  Moreover, the ideas behind SPA are at the heart of many geometric-based near-separable NMF algorithms (see below).

SPA looks for the vertices of the convex hull of the columns of the input data matrix $X$ and works as follows: at each step, it selects the column of $X$ with maximum $\ell_2$ norm and then updates $X$ by projecting each column onto the orthogonal complement of the extracted column; see Algorithm~\ref{spa}. SPA is extremely fast as it can be implemented in $2pnr + \mathcal{O}(pr^2)$ operations, using the formula $||(I-uu^T )v||_2^2 = ||v||_2^2-(u^T v)^2$, for any $u, v \in \mathbb{R}^m$ with $||u||_2 = 1$~\cite{GV12}. Moreover, if $r$ is unknown, it can be estimated using the norm of the residual $R$. 
\renewcommand{\thealgorithm}{SPA}
\algsetup{indent=2em}
\begin{algorithm}[ht!]
\caption{Successive Projection Algorithm~\cite{MC01} \label{spa}}
\begin{algorithmic}[1] 
\REQUIRE Near-separable matrix $\tilde{X} = W[I_r, H']\Pi + N$ where $W$ is full rank, $H' \geq 0$, the entries of each column of $H'$ sum to at most one, $\Pi$ is a permutation and $N$ is the noise, 
and the number~$r$ of columns of~$W$. 
\ENSURE Set of $r$ indices $\mathcal{K}$ such that $\tilde{X}(:,\mathcal{K}) \approx W$ (up to permutation). 
    \medskip  
		
\STATE Let $R = \tilde{X}$, $\mathcal{K} = \{\}$. 
\FOR {$k$ = 1 : $r$}   
\STATE $p = \argmax_j ||R_{:j}||_2$.  
%\STATE $u_k = $.  \vspace{0.1cm} 
\STATE $R = \left(I-\frac{{R_{:p}} R_{:p}^T}{||{R_{:p}}||_2^2}\right)R$. \vspace{0.1cm} 
\STATE $\mathcal{K} = \mathcal{K} \cup \{p\}$. 
\ENDFOR
\end{algorithmic}  
\end{algorithm}  

\begin{figure}[ht!]
\begin{center}
\includegraphics[width=4cm]{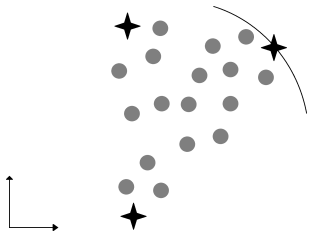}  
\end{center} 
\caption{\small Illustration of SPA.} 
\label{spai}
\end{figure}

Let us prove the correctness of SPA in the noiseless case using induction, and 
assuming $W$ is full rank (this is a necessary and sufficient condition) and 
assuming the entries of each column of $H'$ sum to at most one (this can be achieved through normalization; see above). 
At the first step, SPA identifies a column of $W$ because the $\ell_2$ norm can only be maximized at a vertex 
of the convex hull of a set of points; see Figure~\ref{spai} for an illustration.  
\noindent In fact, for all $1 \leq j \leq n$, 
\[
||X(:,j)||_2 
=  ||WH(:,j)||_2
\leq \sum_{k=1}^r H(k,j) ||W(:,k)||_2 \leq \max_{1 \leq k \leq r} ||W(:,k)||_2. 
\] 
The first inequality follows from the triangle inequality, and the second since $H(k,j) \geq 0$ and $\sum_k H(k,j) \leq 1$. 
Moreover, by strong convexity of the $\ell_2$ norm and the full rank assumption on $W$, the first inequality is strict unless $H(:,k)$ is a column of the identity matrix, that is, unless $X(:,j) = W(:,k)$ for some $k$. 
For the induction step, assume without loss of generality that SPA has extracted the first $\ell$ columns of $W$, 
and let $W_{\ell} = W(:,1$:$\ell)$ and $P_{W_{\ell}}^{\bot}$ be the projection onto the orthogonal complement of the columns of $W_{\ell}$ so that $P_{W_{\ell}}^{\bot} W_{\ell} = 0$. We have, for all $1 \leq j \leq n$,  
\[
||P_{W_{\ell}}^{\bot} X(:,j)||_2 
= ||P_{W_{\ell}}^{\bot}W H(:,j)||_2
\leq  \sum_{k=1}^r H(k,j) ||P_{W_{\ell}}^{\bot} W(:,k)||_2 
%= \sum_{k=1}^r H(k,j) ||P_{W_{\ell}}^{\bot} W(:,k)||_2 
\leq \max_{\ell + 1 \leq k \leq r} ||P_{W_{\ell}}^{\bot} W(:,k)||_2, 
\] 
where $P_{W_{\ell}}^{\bot} W(:,k) \neq 0$ for $\ell + 1 \leq k \leq r$ since $W$ is full rank. 
Hence, using the same reasoning as above, SPA will identify a column of $W$ not extracted yet, which concludes the proof.

Moreover, SPA is robust to noise: given a near-separable matrix $\tilde{X} = W[I_r, H']\Pi + N$ with 
$W$ full rank, 
$H'$ nonnegative with $||H'(:,j)||_1 \leq 1$ $\forall j$, 
and $\epsilon = \max_j ||N(:,j)||_2  \leq \mathcal{O} \left(   \frac{  \sigma_{\min}(W)  }{\sqrt{r} \kappa^2(W)} \right)$, 
 SPA identifies the columns of $W$ up to $\ell_2$ error proportional to $\mathcal{O} \left( \epsilon \, \kappa^2(W) \right)$, 
where $\kappa(W) = \frac{\sigma_{\max}(W)}{\sigma_{\min}(W)}$ \cite[Th.3]{GV12}. 
These bounds can be improved using post-processing (see below) which reduces the error to $\mathcal{O} \left( \epsilon \, \kappa(W) \right)$~\cite{Ar13}, or preconditioning which significantly increases the upper bound on the noise level, to $\epsilon \leq \mathcal{O} \left(  \frac{  \sigma_{\min}(W) }{r \sqrt{r}}\right)$, and reduces the error to $\mathcal{O} \left( \epsilon \, \kappa(W) \right)$ \cite{GV13}. \\

It is interesting to note that SPA has been developed and used for rather different purposes in various fields:  
\begin{itemize}

\item \emph{Numerical linear algebra}. 
SPA is closely related to the modified Gram-Schmidt algorithm with column pivoting, 
used for example to solve linear least squares problems~\cite{BG65}. 

\item \emph{Chemistry} (and in particular spectroscopy). 
SPA is used for variable selection in spectroscopic multicomponent analysis; in fact, the name SPA comes from~\cite{MC01}.

\item \emph{Hyperspectral imaging}. 
SPA is closely related to several endmember extraction algorithms; in particular N-FINDR~\cite{Win99} and its variants, the automatic target generation process (ATGP)~\cite{RC03}, and the successive volume maximization algorithm (SVMAX)~\cite{CM11}; see the discussion in~\cite{Ma14} for more details. The motivation behind all these approaches is to identify an index set $\mathcal{K}$ that maximizes the volume 
%\footnote{That volume can be defined in different ways; one example is the volume of the set \{ x \in \mathbb{R}^{} \ | \ \}} 
of the convex hull of the columns of $X(:,\mathcal{K})$. 
Note that most endmember extraction algorithms use an LDR (such as the SVD) as a pre-processing step for noise filtering, 
and SPA can be combined with an LDR to improve performance. 

\item \emph{Text mining}. Arora et al.~\cite{Ar13} proposed FastAnchorWords whose differences with SPA are that 
(i)~the projection is made onto the affine hull of the columns extracted so far (instead of the linear span), and 
(ii)~the index set extracted is refined using the following post-processing step: let $\mathcal{K}$ be the extracted index set by SPA, 
for each $k \in \mathcal{K}$: 
\begin{itemize}
\item Compute the projection $R$ of $X$ into the orthogonal complement of $X(:, \mathcal{K} \backslash \{k\})$.
\item Replace $k$ with the index corresponding to the column of $R$ with maximum $\ell_2$ norm. 
\end{itemize} 

\item \emph{Theoretical computer science}. 
SPA was proved to be a good heuristic to identify a subset of columns of a given matrix whose convex hull has maximum volume~\cite{AM09, AM10} (in the sense that no polynomial-time algorithm can achieve better performance up to some logarithmic factors). 

\item \emph{Sparse regression with self-dictionary}. SPA is closely related to orthogonal matching pursuit and can be interpreted as a greedy method to solve the sparse regression problem with self-dictionary~\eqref{spreg}; see~\cite{FM13} and the references therein. 

\end{itemize} 

Moreover, there exist many geometric algorithms which are variants of SPA, e.g., 
vertex component analysis (VCA) using linear functions instead of the $\ell_2$-norm~\cite{ND05}, 
$\ell_p$-norm based pure pixel algorithm (TRI-P) using $p$-norms~\cite{A11}, 
FastSepNMF using strongly convex functions~\cite{GV12}, 
the successive nonnegative projection algorithm (SNPA)~\cite{G13} and the fast conical hull algorithm (XRAY)~\cite{KSK12} using nonnegativity constraints for the projection step.  \\

Further research on near-separable NMF includes the design of faster and/or provably more robust algorithms. In particular, there does not seem to exist an algorithm guaranteed to be robust for any matrix $W$ such that $\alpha(W) > 0$ and running in $\mathcal{O}(n)$ operations. 
%It would be an interesting direction for further research to develop both efficient and provably more robust algorithms. 

\section{Connections with Problems in Mathematics and Computer Science}

In this section, we briefly mention several connections of NMF with problems outside data mining and machine learning. 
The minimum $r$ such that an exact NMF of a nonnegative matrix $X$ exists is the nonnegative rank of $X$, denoted $\rank_+(X)$. More precisely, given $X \in \mathbb{R}^{p \times n}_+$, $\rank_+(X)$ is the minimum $r$ such that there exist $W \in \mathbb{R}^{p \times r}_+$ and $H \in \mathbb{R}^{r \times n}_+$ with $X = WH$. 
The nonnegative rank has tight connections with several problems in mathematics and computer science:  
 \begin{itemize}

\item \emph{Graph Theory}. 
Let $G(X)=(V_1 \cup V_2,E)$ be the bipartite graph induced by $X$ (that is, $(i,j) \in E \iff X_{ij} \neq 0$). 
The minimum biclique cover  bc$(G(X))$ of $G(X)$ is the minimum number of complete bipartite subgraphs needed to cover $G(X)$. It can be checked easily that for any $(W,H) \geq 0$ such that $X = WH = \sum_{k=1}^r W_{:k} H_{k:}$, we have
\[
G(X) = \cup_{k=1}^r G(W_{:k} H_{k:}) , 
\] 
where $G(W_{:k} H_{k:})$ are complete bipartite subgraphs hence bc$(G(W_{:k} H_{k:})) = 1$ $\forall k$. Therefore, 
\[
\text{bc}(G(X)) \leq \rank_+(X) .  
\]
This lower bound on the nonnegative rank is referred to as the rectangle covering bound \cite{FK11}.

\item \emph{Extended Formulations}. Given a polytope $P$, an extended formulation (or lift, or extension) is a higher dimensional polyhedron $Q$ and a linear projection $\pi$ such that $\pi(Q) = P$. When the polytope $P$ has exponentially many facets, finding extended formulations of polynomial size is of great importance since it allows to solve linear programs (LP) over $P$ in polynomial time. 
It turns out that the minimum number of facets of an extended formulation $Q$ of a polytope $P$ is equal to the nonnegative rank of its slack matrix \cite{Y91}, defined as $X(i,j) = a_i^T v_j - b_i$ where $v_j$ is the $j$th vertex of $P$ and \mbox{$\{ x \in \mathbb{R}^n \ | \ a_i^T x - b_i \geq 0 \}$} its $i$th facet with $a_i \in \mathbb{R}^n$ and $b_i \in \mathbb{R}$, that is, $X$ is a facet-by-vertex matrix and $X(i,j)$ is the slack of the  $j$th vertex with respect to $i$th facet; see the surveys \cite{CCCZ09, K11} and the references therein.  
These ideas can be generalized to approximate extended formulations, directly related to approximate factorizations (hence NMF) \cite{BFK12, Gouv13}.

\item \emph{Probability}. Let $X^{(k)} \in \{1,\dots,p\}$ and $Y^{(k)} \in \{1,\dots,n\}$  be two independent variables for each $1 \leq k \leq r$, and $P^{(k)}$ be the joint distribution with 
\[ 
P_{ij}^{(k)} 
= \mathbb{P}\left(X^{(k)}=i,Y^{(k)}=j\right) 
= \mathbb{P}\left(X^{(k)}=i\right)  \mathbb{P}\left(Y^{(k)}=j\right) . 
\]
Each distribution $P^{(k)}$ corresponds to a nonnegative rank-one matrix. 
Let us define the mixture $P$ of these $k$ independent distributions as follows: 
\begin{itemize}
\item Choose the distribution $P^{(k)}$ with probability $\alpha_{k}$, where $\sum_{k = 1}^r \alpha_{k} = 1$. 
\item Draw $X$ and $Y$ from the distribution $P^{(k)}$.  
\end{itemize}
We have that $P = \sum_{k = 1}^r \alpha_{k} P^{(k)}$ is the sum of $r$ rank-one nonnegative matrices. %Therefore, given the mixture $P$ of independent models, 
%Recovering the $k$ original distributions and the corresponding mixing coefficients $\alpha_l$ from the distribution $P$ is equivalent to computing a nonnegative factorization of $P$. 
In practice, only $P$ is observed and computing its nonnegative rank and a corresponding factorization amounts to explaining the distribution $P$ with as few independent variables as possible; see \cite{CR10} and the references therein.

\item \emph{Communication Complexity}. In its simplest variant, communication complexity addresses the following problem: Alice and Bob have to compute the following function 
\[
f : \{0,1\}^{m} \times \{0,1\}^{n}  \mapsto  \{0,1\} : (x,y) \mapsto f(x,y).  
\]
Alice only knows $x$ and Bob $y$, and the aim is to minimize the number of bits exchanged between Alice and Bob to compute $f$ exactly. 
 Nondeterministic communication complexity (NCC) is a variant 
%of communication complexity 
where Bob and Alice first receive a message before starting their communication; 
see \cite[Ch.3]{Lee09} and the references therein for more details. 
The communication matrix $X \in \{0,1\}^{2^n \times 2^m}$ is equal to the function $f$ for all possible combinations of inputs. 
Yannakakis \cite{Y91} showed that the NCC for computing $f$ is upper bounded by the logarithm of the nonnegative rank of the communication matrix (this result is closely related to the rectangle covering bound described above: in fact, $\left\lceil  \log(\text{bc}(G(X)) \right\rceil$ equals to the NCC of $f$).

\item \emph{Computational Geometry}. Computing the nonnegative rank is closely related to the problem of finding a polytope with minimum number of vertices nested between two given polytopes \cite{GG10b}. This is a well-known problem is computational geometry, referred to as the nested polytopes problem; see \cite{DJ90} and the references therein.

\end{itemize}

\section{Conclusion}

\begin{figure}[ht!]
\begin{center}
\includegraphics[width=7cm]{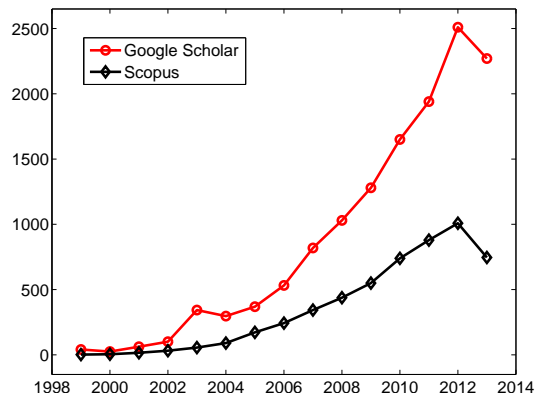} 
\end{center} 
\caption{\small Number of search results for papers containing either `nonnegative matrix factorization' or `non-negative matrix factorization' on Google Scholar and Scopus (as of December 12, 2013).}  
\label{nmfdata}
\end{figure}

NMF is an easily interpretable linear dimensionality reduction technique for nonnegative data. 
It is a rather versatile technique with many applications, and brings together a broad range of researchers. 
In the context of `Big Data' science, which becomes an increasingly important topic, we believe NMF has a bright future; see Figure~\ref{nmfdata} for an illustration of the number of publications related to NMF since the publication of the Lee and Seung paper \cite{LS99}.  

\section*{Acknowledgment} 

The author would like to thank Rafal Zdunek, Wing-Kin Ma, Marc Pirlot and the Editors of the book `Regularization, Optimization, Kernels, and
Support Vector Machines' for insightful comments which helped improve the paper.

\small        

\bibliographystyle{spmpsci}
\bibliography{Biography}

\begin{thebibliography}{100}
\providecommand{\url}[1]{{#1}}
\providecommand{\urlprefix}{URL }
\expandafter\ifx\csname urlstyle\endcsname\relax
  \providecommand{\doi}[1]{DOI~\discretionary{}{}{}#1}\else
  \providecommand{\doi}{DOI~\discretionary{}{}{}\begingroup
  \urlstyle{rm}\Url}\fi

\bibitem{A11}
Ambikapathi, A., Chan, T.H., Chi, C.Y., Keizer, K.: Hyperspectral data geometry
  based estimation of number of endmembers using p-norm based pure pixel
  identification.
\newblock IEEE Trans. on Geoscience and Remote Sensing \textbf{51}(5),
  2753--2769 (2013)

\bibitem{MC01}
Ara\'ujo, U., Saldanha, B., Galv\~ao, R., Yoneyama, T., Chame, H., Visani, V.:
  The successive projections algorithm for variable selection in spectroscopic
  multicomponent analysis.
\newblock Chemometrics and Intelligent Laboratory Systems \textbf{57}(2),
  65--73 (2001)

\bibitem{Ar13}
Arora, S., Ge, R., Halpern, Y., Mimno, D., Moitra, A., Sontag, D., Wu, Y., Zhu,
  M.: A practical algorithm for topic modeling with provable guarantees.
\newblock In: Int. Conf. on Machine Learning (ICML~'13), vol.~28, pp. 280--288
  (2013)

\bibitem{AGKM11}
Arora, S., Ge, R., Kannan, R., Moitra, A.: Computing a nonnegative matrix
  factorization -- provably.
\newblock In: Proc. of the 44th Symp. on Theory of Computing (STOC~'12), pp.
  145--162 (2012)

\bibitem{BBV}
Badeau, R., Bertin, N., Vincent, E.: Stability analysis of multiplicative
  update algorithms and application to nonnegative matrix factorization.
\newblock IEEE Trans. on Neural Networks \textbf{21}(12), 1869--1881 (2010)

\bibitem{BBLPP07}
Berry, M., Browne, M., Langville, A., Pauca, V., Plemmons, R.: {Algorithms and
  Applications for Approximate Nonnegative Matrix Factorization}.
\newblock Computational Statistics \& Data Analysis \textbf{52}, 155--173
  (2007)

\bibitem{BN05}
Bioucas-Dias, J., Nascimento, J.: Estimation of signal subspace on
  hyperspectral data.
\newblock In: Remote Sensing, p. 59820L. International Society for Optics and
  Photonics (2005)

\bibitem{BP12}
Bioucas-Dias, J., Plaza, A., Dobigeon, N., Parente, M., Du, Q., Gader, P.,
  Chanussot, J.: Hyperspectral unmixing overview: Geometrical, statistical, and
  sparse regression-based approaches.
\newblock IEEE J. of Selected Topics in Applied Earth Observations and Remote
  Sensing \textbf{5}(2), 354--379 (2012)

\bibitem{BRRT12}
Bittorf, V., Recht, B., R\'{e}, E., Tropp, J.: {Factoring nonnegative matrices
  with linear programs}.
\newblock In: Advances in Neural Information Processing Systems (NIPS~'12), pp.
  1223--1231 (2012)

\bibitem{Blei12}
Blei, D.: Probabilistic topic models.
\newblock Communications of the ACM \textbf{55}(4), 77--84 (2012)

\bibitem{Bou}
Boutsidis, C., Gallopoulos, E.: {SVD based initialization: A head start for
  nonnegative matrix factorization}.
\newblock Pattern Recognition \textbf{41}, 1350--1362 (2008)

\bibitem{BM09}
Boutsidis, C., Mahoney, M., Drineas, P.: An improved approximation algorithm
  for the column subset selection problem.
\newblock In: Proc. of the 20th Annual ACM-SIAM Symp. on Discrete Algorithms
  (SODA~'09), pp. 968--977 (2009)

\bibitem{BFK12}
Braun, G., Fiorini, S., Pokutta, S., Steurer, D.: Approximation limits of
  linear programs (beyond hierarchies).
\newblock In: Proc. of the 53rd Annual IEEE Symp. on Foundations of Computer
  Science (FOCS'~12), pp. 480--489 (2012)

\bibitem{Bro98}
Bro, R.: Multi-way analysis in the food industry: Models, algorithms, and
  applications.
\newblock Ph.D. thesis, University of Copenhagen (1998).
\newblock \url{http://curis.ku.dk/ws/files/13035961/Rasmus_Bro.pdf}

\bibitem{BG65}
Businger, P., Golub, G.: Linear least squares solutions by householder
  transformations.
\newblock Numerische Mathematik \textbf{7}, 269--276 (1965)

\bibitem{CH11}
Cai, D., He, X., Han, J., Huang, T.: Graph regularized nonnegative matrix
  factorization for data representation.
\newblock IEEE Trans. on Pattern Analysis and Machine Intelligence
  \textbf{33}(8), 1548--1560 (2011)

\bibitem{CR10}
Carlini, E., Rapallo, F.: {Probability matrices, non-negative rank, and
  parameterization of mixture models}.
\newblock Linear Algebra and its Applications \textbf{433}, 424--432 (2010)

\bibitem{Casa13}
Casalino, G., Del~Buono, N., Mencar, C.: Subtractive clustering for seeding
  non-negative matrix factorizations.
\newblock Information Sciences (257), 369--387 (2013)

\bibitem{AM09}
\c{C}ivril, A., Magdon-Ismail, M.: On selecting a maximum volume sub-matrix of
  a matrix and related problems.
\newblock Theoretical Computer Science \textbf{410}(47-49), 4801--4811 (2009)

\bibitem{AM10}
\c{C}ivril, A., Magdon-Ismail, M.: Exponential inapproximability of selecting a
  maximum volume sub-matrix.
\newblock Algorithmica \textbf{65}(1), 159--176 (2013)

\bibitem{CM11}
Chan, T.H., Ma, W.K., Ambikapathi, A., Chi, C.Y.: A simplex volume maximization
  framework for hyperspectral endmember extraction.
\newblock IEEE Trans. on Geoscience and Remote Sensing \textbf{49}(11),
  4177--4193 (2011)

\bibitem{CMCW08}
Chan, T.H., Ma, W.K., Chi, C.Y., Wang, Y.: A convex analysis framework for
  blind separation of non-negative sources.
\newblock IEEE Trans. on Signal Processing \textbf{56}(10), 5120--5134 (2008)

\bibitem{CK12}
Chi, E., Kolda, T.: On tensors, sparsity, and nonnegative factorizations.
\newblock SIAM J. on Matrix Analysis and Applications \textbf{33}(4),
  1272--1299 (2012)

\bibitem{Ch08}
Choi, S.: Algorithms for orthogonal nonnegative matrix factorization.
\newblock In: Proc.\@ of the Int.\@ Joint Conf.\@ on Neural Networks, pp.
  1828--1832 (2008)

\bibitem{CP09b}
Cichocki, A., Phan, A.: {Fast local algorithms for large scale Nonnegative
  Matrix and Tensor Factorizations}.
\newblock IEICE Trans. on Fundamentals of Electronics \textbf{Vol. E92-A No.3},
  708--721 (2009)

\bibitem{CZA06}
Cichocki, A., Zdunek, R., Amari, S.I.: {Non-negative Matrix Factorization with
  Quasi-Newton Optimization}.
\newblock In: Lecture Notes in Artificial Intelligence, Springer, vol. 4029,
  pp. 870--879 (2006)

\bibitem{CZA07}
Cichocki, A., Zdunek, R., Amari, S.I.: {Hierarchical ALS Algorithms for
  Nonnegative Matrix and 3D Tensor Factorization}.
\newblock In: Lecture Notes in Computer Science, Vol. 4666, Springer, pp.
  169--176 (2007)

\bibitem{CAZP09}
Cichocki, A., Zdunek, R., Phan, A., Amari, S.I.: Nonnegative Matrix and Tensor
  Factorizations: Applications to Exploratory Multi-way Data Analysis and Blind
  Source Separation.
\newblock Wiley-Blackwell (2009)

\bibitem{C94}
Comon, P.: {Independent component analysis, A new concept?}
\newblock Signal Processing \textbf{36}, 287--314 (1994)

\bibitem{CCCZ09}
Conforti, M., Cornu\'ejols, G., Zambelli, G.: Extended formulations in
  combinatorial optimization.
\newblock 4OR: A Quarterly Journal of Operations Research \textbf{10}(1), 1--48
  (2010)

\bibitem{DJ90}
Das, G., Joseph, D.: {The Complexity of Minimum Convex Nested Polyhedra}.
\newblock In: Proc. of the 2nd Canadian Conf. on Computational Geometry, pp.
  296--301 (1990)

\bibitem{AE07}
d'Aspremont, A., El~Ghaoui, L., Jordan, M., Lanckriet, G.: {A Direct
  Formulation for Sparse PCA Using Semidefinite Programming}.
\newblock SIAM Review \textbf{49}(3), 434--448 (2007)

\bibitem{DM86}
Daube-Witherspoon, M., Muehllehner, G.: An iterative image space reconstruction
  algorithm suitable for volume {ECT}.
\newblock IEEE Trans. on Medical Imaging \textbf{5}, 61--66 (1986)

\bibitem{D08}
Devarajan, K.: {Nonnegative Matrix Factorization: An Analytical and
  Interpretive Tool in Computational Biology}.
\newblock PLoS Computational Biology \textbf{4}(7), {e1000029} (2008)

\bibitem{DHS05}
Ding, C., He, X., Simon, H.: {On the Equivalence of Nonnegative Matrix
  Factorization and Spectral Clustering}.
\newblock In: SIAM Int. Conf. Data Mining (SDM'05), pp. 606--610 (2005)

\bibitem{DJ10}
Ding, C., Li, T., Jordan, M.: Convex and semi-nonnegative matrix
  factorizations.
\newblock IEEE Trans. on Pattern Analysis and Machine Intelligence
  \textbf{32}(1), 45--55 (2010)

\bibitem{DLTP08}
Ding, C., Li, T., Peng, W.: On the equivalence between non-negative matrix
  factorization and probabilistic latent semantic indexing.
\newblock Computational Statistics \& Data Analysis \textbf{52}(8), 3913--3927
  (2008)

\bibitem{DH06}
Ding, C., Li, T., Peng, W., Park, H.: Orthogonal nonnegative matrix
  t-factorizations for clustering.
\newblock In: Proc. of the 12th ACM SIGKDD Int. Conf. on Knowledge Discovery
  and Data Mining, pp. 126--135 (2006)

\bibitem{DRIS13}
Ding, W., Rohban, M., Ishwar, P., Saligrama, V.: Topic discovery through data
  dependent and random projections.
\newblock In: Int. Conf. on Machine Learning (ICML~'13), vol.~28, pp. 471--479
  (2013)

\bibitem{ESV12}
Elhamifar, E., Sapiro, G., Vidal, R.: See all by looking at a few: Sparse
  modeling for finding representative objects.
\newblock In: IEEE Conf. on Computer Vision and Pattern Recognition (CVPR~'12)
  (2012)

\bibitem{EMO12}
Esser, E., Moller, M., Osher, S., Sapiro, G., Xin, J.: A convex model for
  nonnegative matrix factorization and dimensionality reduction on physical
  space.
\newblock IEEE Trans. on Image Processing \textbf{21}(7), 3239--3252 (2012)

\bibitem{FCB09}
F{\'e}votte, C., Bertin, N., Durrieu, J.L.: Nonnegative matrix factorization
  with the {I}takura-{S}aito divergence: {W}ith application to music analysis.
\newblock Neural Computation \textbf{21}(3), 793--830 (2009)

\bibitem{FK11}
Fiorini, S., Kaibel, V., Pashkovich, K., Theis, D.: Combinatorial bounds on
  nonnegative rank and extended formulations.
\newblock Discrete Mathematics \textbf{313}(1), 67--83 (2013)

\bibitem{FM13}
Fu, X., Ma, W.K., Chan, T.H., Bioucas-Dias, J., Iordache, M.D.: Greedy
  algorithms for pure pixels identification in hyperspectral unmixing: A
  multiple-measurement vector viewpoint.
\newblock In: Proc. of 21st European Signal Processing Conf. (EUSIPCO~'13)
  (2013)

\bibitem{GG05}
Gaussier, E., Goutte, C.: Relation between {PLSA} and {NMF} and implications.
\newblock In: Proc. of the 28th Annual Int. ACM SIGIR Conf. on Research and
  Development in Information Retrieval, pp. 601--602 (2005)

\bibitem{Ge13}
Ge, R.: Provable algorithms for machine learning problems.
\newblock Ph.D. thesis, Princeton University (2013).
\newblock
  \url{http://dataspace.princeton.edu/jspui/bitstream/88435/dsp019k41zd62n/1/Ge_princeton_0181D_10819.pdf}

\bibitem{NG11}
Gillis, N.: Nonnegative matrix factorization: Complexity, algorithms and
  applications.
\newblock Ph.D. thesis, Universit\'{e} catholique de Louvain (2011).
\newblock \url{https://sites.google.com/site/nicolasgillis/}

\bibitem{G12}
Gillis, N.: {Sparse and unique nonnegative matrix factorization through data
  preprocessing}.
\newblock Journal of Machine Learning Research \textbf{13}(Nov), 3349--3386
  (2012)

\bibitem{G13h}
Gillis, N.: Robustness analysis of {Hottopixx}, a linear programming model for
  factoring nonnegative matrices.
\newblock SIAM J. on Matrix Analysis and Applications \textbf{34}(3),
  1189--1212 (2013)

\bibitem{G13}
Gillis, N.: Successive nonnegative projection algorithm for robust nonnegative
  blind source separation (2013).
\newblock {arXiv:1310.7529}

\bibitem{GG09}
Gillis, N., Glineur, F.: Using underapproximations for sparse nonnegative
  matrix factorization.
\newblock Pattern Recognition \textbf{43}(4), 1676--1687 (2010)

\bibitem{GG12}
Gillis, N., Glineur, F.: Accelerated multiplicative updates and hierarchical
  \textsc{ALS} algorithms for nonnegative matrix factorization.
\newblock Neural Computation \textbf{24}(4), 1085--1105 (2012)

\bibitem{GG10b}
Gillis, N., Glineur, F.: On the geometric interpretation of the nonnegative
  rank.
\newblock Linear Algebra and its Applications \textbf{437}(11), 2685--2712
  (2012)

\bibitem{GL13}
Gillis, N., Luce, R.: Robust near-separable nonnegative matrix factorization
  using linear optimization.
\newblock Journal of Machine Learning Research  (2014).
\newblock {to appear}

\bibitem{GV12}
Gillis, N., Vavasis, S.: Fast and robust recursive algorithms for separable
  nonnegative matrix factorization.
\newblock IEEE Trans. Pattern Anal. Mach. Intell.  (2013).
\newblock {doi:10.1109/TPAMI.2013.226}

\bibitem{GV13}
Gillis, N., Vavasis, S.: Semidefinite programming based preconditioning for
  more robust near-separable nonnegative matrix factorization (2013).
\newblock {arXiv:1310.2273}

\bibitem{GV96}
Golub, G., Van~Loan, C.: Matrix Computation, 3rd Edition.
\newblock The Johns Hopkins University Press Baltimore (1996)

\bibitem{Gouv13}
Gouveia, J., Parrilo, P., Thomas, R.: Approximate cone factorizations and lifts
  of polytopes (2013).
\newblock {arXiv:1308.2162}

\bibitem{GS00}
Grippo, L., Sciandrone, M.: {On the convergence of the block nonlinear
  Gauss-Seidel method under convex constraints}.
\newblock Operations Research Letters \textbf{26}, 127--136 (2000)

\bibitem{GTA12}
Guan, N., Tao, D., Luo, Z., Yuan, B.: {NeNMF: an optimal gradient method for
  nonnegative matrix factorization}.
\newblock IEEE Trans. on Signal Processing \textbf{60}(6), 2882--2898 (2012)

\bibitem{GV02}
Guillamet, D., Vitri{\`a}, J.: Non-negative matrix factorization for face
  recognition.
\newblock In: Lecture Notes in Artificial Intelligence, pp. 336--344. Springer
  (2002)

\bibitem{Han09}
Han, J., Han, L., Neumann, M., Prasad, U.: On the rate of convergence of the
  image space reconstruction algorithm.
\newblock Operators and Matrices \textbf{3}(1), 41--58 (2009)

\bibitem{Ho08}
Ho, N.D.: Nonnegative matrix factorization - algorithms and applications.
\newblock Ph.D. thesis, Universit\'{e} catholique de Louvain (2008)

\bibitem{Hoy}
Hoyer, P.: Nonnegative matrix factorization with sparseness constraints.
\newblock Journal of Machine Learning Research \textbf{5}, 1457--1469 (2004)

\bibitem{HD11}
Hsieh, C.J., Dhillon, I.: Fast coordinate descent methods with variable
  selection for non-negative matrix factorization.
\newblock In: Proc. of the 17th ACM SIGKDD int. conf. on Knowledge discovery
  and data mining, pp. 1064--1072 (2011)

\bibitem{HSS14}
Huang, K., Sidiropoulos, N., Swami, A.: Non-negative matrix factorization
  revisited: {U}niqueness and algorithm for symmetric decomposition.
\newblock IEEE Trans. on Signal Processing \textbf{62}(1), 211--224 (2014)

\bibitem{IBP12}
Iordache, M.D., Bioucas-Dias, J., Plaza, A.: Total variation spatial
  regularization for sparse hyperspectral unmixing.
\newblock IEEE Trans. on Geoscience and Remote Sensing \textbf{50}(11),
  4484--4502 (2012)

\bibitem{JQ09}
Jia, S., Qian, Y.: Constrained nonnegative matrix factorization for
  hyperspectral unmixing.
\newblock IEEE Trans.\@ on Geoscience and Remote Sensing \textbf{47}(1),
  161--173 (2009)

\bibitem{K11}
Kaibel, V.: {Extended Formulations in Combinatorial Optimization}.
\newblock Optima \textbf{85}, 2--7 (2011)

\bibitem{KS10}
Kanagal, B., Sindhwani, V.: Rank selection in low-rank matrix approximations.
\newblock In: Advances in Neural Information Processing Systems (NIPS~'10)
  (2010)

\bibitem{KK05}
Ke, Q., Kanade, T.: Robust {L}$_1$ norm factorization in the presence of
  outliers and missing data by alternative convex programming.
\newblock In: IEEE Conf. on Computer Vision and Pattern Recognition (CVPR~'05),
  pp. 739--746 (2005)

\bibitem{KP07}
Kim, H., Park, H.: {Sparse non-negative matrix factorizations via alternating
  non-negativity-constrained least squares for microarray data analysis}.
\newblock Bioinformatics \textbf{23}(12), 1495--1502 (2007)

\bibitem{KP08}
Kim, H., Park, H.: {Non-negative Matrix Factorization Based on Alternating
  Non-negativity Constrained Least Squares and Active Set Method}.
\newblock SIAM J. on Matrix Analysis and Applications \textbf{30}(2), 713--730
  (2008)

\bibitem{KHP13}
Kim, J., He, Y., Park, H.: Algorithms for nonnegative matrix and tensor
  factorizations: {A} unified view based on block coordinate descent framework.
\newblock Journal of Global Optimization  (2013).
\newblock {doi:10.1007/s10898-013-0035-4}

\bibitem{KP11}
Kim, J., Park, H.: Fast nonnegative matrix factorization: An active-set-like
  method and comparisons.
\newblock SIAM J. on Scientific Computing \textbf{33}(6), 3261--3281 (2011)

\bibitem{KS13}
Kumar, A., Sindhwani, V.: Near-separable non-negative matrix factorization with
  $\ell_1$- and {B}regman loss functions (2013).
\newblock {arXiv:1312.7167}

\bibitem{KSK12}
Kumar, A., Sindhwani, V., Kambadur, P.: Fast conical hull algorithms for
  near-separable non-negative matrix factorization.
\newblock In: Int. Conf. on Machine Learning (ICML~'13), vol.~28, pp. 231--239
  (2013)

\bibitem{LH74}
Lawson, C., Hanson, R.: Solving Least Squares Problems.
\newblock Prentice-Hall (1974)

\bibitem{LS99}
Lee, D., Seung, H.: {Learning the Parts of Objects by Nonnegative Matrix
  Factorization}.
\newblock Nature \textbf{401}, 788--791 (1999)

\bibitem{LS01}
Lee, D., Seung, H.: {Algorithms for Non-negative Matrix Factorization}.
\newblock In Advances in Neural Information Processing (NIPS~'01) \textbf{13}
  (2001)

\bibitem{Lee09}
Lee, T., Shraibman, A.: Lower bounds in communication complexity.
\newblock Now Publishers Inc. (2009)

\bibitem{AL11}
Lef\`evre, A.: Dictionary learning methods for single-channel source
  separations.
\newblock Ph.D. thesis, Ecole Normale Sup\'erieure de Cachan (2012)

\bibitem{LLP12}
Li, L., Lebanon, G., Park, H.: Fast {B}regman divergence {NMF} using {T}aylor
  expansion and coordinate descent.
\newblock In: Proc. of the 18th ACM SIGKDD Int. Conf. on Knowledge Discovery
  and Data Mining, pp. 307--315 (2012)

\bibitem{LZ09}
Li, L., Zhang, Y.J.: {FastNMF: highly efficient monotonic fixed-point
  nonnegative matrix factorization algorithm with good applicability}.
\newblock J. Electron. Imaging \textbf{Vol. 18}(033004) (2009)

\bibitem{LZS09}
Li, T., Zhang, Y., Sindhwani, V.: A non-negative matrix tri-factorization
  approach to sentiment classification with lexical prior knowledge.
\newblock In: Association of Computational Lingustics, pp. 244--252 (2009)

\bibitem{NBM13}
Li, Y., Sima, D., Van~Cauter, S., Croitor~Sava, A., Himmelreich, U., Pi, Y.,
  Van~Huffel, S.: {Hierarchical non-negative matrix factorization (hNMF): a
  tissue pattern differentiation method for glioblastoma multiforme diagnosis
  using MRSI}.
\newblock NMR in Biomedicine \textbf{26}(3), 307--319 (2013)

\bibitem{Lin07}
Lin, C.J.: On the convergence of multiplicative update algorithms for
  nonnegative matrix factorization.
\newblock IEEE Trans. on Neural Networks \textbf{18}(6), 1589--1596 (2007)

\bibitem{L07}
Lin, C.J.: {Projected Gradient Methods for Nonnegative Matrix Factorization}.
\newblock Neural Computation \textbf{19}, 2756--2779 (2007)

\bibitem{LW12}
Liu, J., Liu, J., Wonka, P., Ye, J.: Sparse non-negative tensor factorization
  using columnwise coordinate descent.
\newblock Pattern Recognition \textbf{45}(1), 649--656 (2012)

\bibitem{Ma14}
Ma, W.K., Bioucas-Dias, J., Chan, T.H., Gillis, N., Gader, P., Plaza, A.,
  Ambikapathi, A., Chi, C.Y.: {A Signal Processing Perspective on Hyperspectral
  Unmixing}.
\newblock IEEE Signal Processing Magazine \textbf{31}(1), 67--81 (2014)

\bibitem{MD09}
Mahoney, M., Drineas, P.: {CUR matrix decompositions for improved data
  analysis}.
\newblock Proc. of the National Academy of Sciences \textbf{106}(3), 697--702
  (2009)

\bibitem{Vikas10}
Melville, P., Sindhwani, V.: Recommender systems.
\newblock Encyclopedia of machine learning \textbf{1}, 829--838 (2010)

\bibitem{MH07}
Miao, L., Qi, H.: Endmember extraction from highly mixed data using minimum
  volume constrained nonnegative matrix factorization.
\newblock IEEE Trans. on Geoscience and Remote Sensing \textbf{45}(3), 765--777
  (2007)

\bibitem{Moit13}
Moitra, A.: An almost optimal algorithm for computing nonnegative rank.
\newblock In: Proc. of the 24th Annual ACM-SIAM Symp. on Discrete Algorithms
  (SODA~'13), pp. 1454--1464 (2013)

\bibitem{NN05}
Naanaa, W., Nuzillard, J.M.: Blind source separation of positive and partially
  correlated data.
\newblock Signal Processing \textbf{85}(9), 1711--1722 (2005)

\bibitem{ND05}
Nascimento, J., Bioucas-Dias, J.: Vertex component analysis: a fast algorithm
  to unmix hyperspectral data.
\newblock IEEE Trans. on Geoscience and Remote Sensing \textbf{43}(4), 898--910
  (2005)

\bibitem{PT94}
Paatero, P., Tapper, U.: {Positive matrix factorization: a non-negative factor
  model with optimal utilization of error estimates of data values}.
\newblock Environmetrics \textbf{5}, 111--126 (1994)

\bibitem{RBL13}
Rapin, J., Bobin, J., Larue, A., Starck, J.L.: Sparse and non-negative {BSS}
  for noisy data.
\newblock IEEE Trans. on Signal Processing \textbf{61}(22), 5620--5632 (2013)

\bibitem{RC03}
Ren, H., Chang, C.I.: Automatic spectral target recognition in hyperspectral
  imagery.
\newblock IEEE Trans. on Aerospace and Electronic Systems \textbf{39}(4),
  1232--1249 (2003)

\bibitem{SL09}
Sandler, R., Lindenbaum, M.: Nonnegative matrix factorization with earth
  mover's distance metric.
\newblock In: IEEE Conf. on Computer Vision and Pattern Recognition (CVPR~'09),
  pp. 1873--1880 (2009)

\bibitem{SBPP06}
Shahnaz, F., Berry, M., A., Pauca, V., Plemmons, R.: {Document clustering using
  nonnegative matrix factorization}.
\newblock Information Processing and Management \textbf{42}, 373--386 (2006)

\bibitem{SFMM14}
Smaragdis, P., F\'evotte, C., Mysore, G., Mohammadiha, N., Hoffman, M.: {A
  Uniﬁed View of Static and Dynamic Source Separation Using Non-Negative
  Factorizations}.
\newblock IEEE Signal Processing Magazine  (2014)

\bibitem{TR13}
Takahashi, N., Hibi, R.: Global convergence of modified multiplicative updates
  for nonnegative matrix factorization.
\newblock Computational Optimization and Applications  (2013).
\newblock {doi:10.1007/s10589-013-9593-0}

\bibitem{TF09}
Tan, V., F{\'e}votte, C.: Automatic relevance determination in nonnegative
  matrix factorization.
\newblock In: Signal Processing with Adaptive Sparse Structured Representations
  (SPARS~'09) (2009)

\bibitem{V09}
Vavasis, S.: On the complexity of nonnegative matrix factorization.
\newblock SIAM J. on Optimization \textbf{20}(3), 1364--1377 (2009)

\bibitem{WL11}
Wang, F., Li, T., Wang, X., Zhu, S., Ding, C.: Community discovery using
  nonnegative matrix factorization.
\newblock Data Min. Knowl. Disc. \textbf{22}(3), 493--521 (2011)

\bibitem{Wild04}
Wild, S., Curry, J., Dougherty, A.: Improving non-negative matrix
  factorizations through structured initialization.
\newblock Pattern Recognition \textbf{37}(11), 2217--2232 (2004)

\bibitem{Win99}
Winter, M.: {N-FINDR: an algorithm for fast autonomous spectral end-member
  determination in hyperspectral data}.
\newblock In: Proc. SPIE Conf. on Imaging Spectrometry V, pp. 266--275 (1999)

\bibitem{X08}
Xue, Y., Tong, C., Chen, Y., Chen, W.S.: Clustering-based initialization for
  non-negative matrix factorization.
\newblock Applied Mathematics and Computation \textbf{205}(2), 525--536 (2008)

\bibitem{YO10}
Yang, Z., Oja, E.: Linear and nonlinear projective nonnegative matrix
  factorization.
\newblock IEEE Trans. on Neural Networks \textbf{21}(5), 734--749 (2010)

\bibitem{Y91}
Yannakakis, M.: {Expressing Combinatorial Optimization Problems by Linear
  Programs}.
\newblock Journal of Computer and System Sciences \textbf{43}(3), 441--466
  (1991)

\bibitem{Zd12}
Zdunek, R.: Initialization of nonnegative matrix factorization with vertices of
  convex polytope.
\newblock In: Artificial Intelligence and Soft Computing, \emph{Lecture Notes
  in Computer Science}, vol. 7267, pp. 448--455. Springer Berlin Heidelberg
  (2012)

\bibitem{ZG05}
Zhong, S., Ghosh, J.: Generative model-based document clustering: a comparative
  study.
\newblock Knowledge and Information Systems \textbf{8}(3), 374--384 (2005)

\end{thebibliography}

\end{document}